\documentclass[11pt]{article}
\usepackage{hyperref}
\usepackage{amsmath, amsfonts, amssymb, mathrsfs}
\usepackage{dcolumn}
\usepackage{caption}
\usepackage{subcaption}
\usepackage{filemod}
\usepackage[ruled, vlined]{algorithm2e}
\usepackage{floatrow}
\usepackage{setspace}
\usepackage{graphicx}
\usepackage[toc,page]{appendix}
\graphicspath{{./Figures/}}

\usepackage{natbib}

\DeclareMathOperator*{\argmin}{\arg\!\min}
\DeclareMathOperator*{\argmax}{\arg\!\max}

\newcommand{\blu}[1]{\textcolor{black}{#1}} 

\title{\vspace{-1cm}Robust expected improvement for Bayesian optimization}
\author{Ryan B.~Christianson\thanks{Corresponding author: Department of 
Statistics and Data Science, NORC at the University of Chicago, {\tt 
christianson-ryan@norc.org}} \and 
Robert B.~Gramacy\thanks{Department of Statistics, Virginia Tech} }
\date{\today}

\begin{document}
\maketitle

\begin{abstract}
Bayesian Optimization (BO) links Gaussian Process (GP) surrogates with
sequential design toward optimizing expensive-to-evaluate black-box
functions. Example design heuristics, or so-called acquisition functions,
like expected improvement (EI), balance exploration and exploitation to
furnish global solutions under stringent evaluation budgets. However, they
fall short when solving for robust optima, meaning a preference for
solutions in a wider domain of attraction. 
Robust solutions are useful when inputs are imprecisely specified, or where a
series of solutions is desired. 
A common mathematical programming technique in such settings involves an adversarial objective, biasing a local solver away from
``sharp'' troughs. Here we propose a surrogate modeling and active
learning technique called robust expected improvement (REI) that ports
adversarial methodology into the BO/GP framework.  
After describing the methods, we illustrate and draw comparisons to several
competitors on benchmark synthetic \blu{exercises} and real problems of varying
complexity.
\end{abstract}



\section{Introduction}
\label{sec:intro}

Globally optimizing a black-box function $f$, finding
\begin{equation}
	\label{eq:bogoal}
	x^* = \argmin_{x \in \mathcal [0,1]^d} \ f(x),
\end{equation}
is a common problem in recommender systems \citep{Vanchinathan2014},
hyperparameter tuning \citep{Snoek2012}, inventory management
\citep{Hong2006}, and engineering \citep{Randall1981}.  Here we explore a
robot pushing problem \citep{Kaelbling2017}. In such settings, $f$ is an
expensive to evaluate computer simulation, so one must carefully design an
experiment to effectively learn the function \citep{Sacks1989} and isolate its
local or global minima. Optimization via modeling and design has a rich
history in statistics \citep{box2007response,myers2016response}. Its modern
instantiation is known as Bayesian optimization
\citep[BO;][]{mockus1978application}.

In BO, one fits a flexible response surface to a limited campaign of
example runs, obtaining a so-called {\em surrogate} $\hat{f}$
\citep{Gramacy2020}.  Based on that fit
-- and in particular its predictive equations for new locations $x$ -- one
   then devises a criteria, a so-called \emph{acquisition function}
   \citep{Shahriari2016}, targeting desirable qualities, such as $x^\star$
   that minimize $f$.  One must choose a surrogate family for $f$, pair it
   with a fitting scheme for $\hat{f}$, and choose a criteria to solve for
   acquisitions.   
   BO is an example
   of {\em active learning (AL)} where one attempts to \blu{create} a virtuous
   cycle of learning and data collection. Many good solutions exist in this
   context, and we shall not provide an in-depth review here.

Perhaps the most common surrogate for BO 
is the Gaussian Process
\citep[GP;][]{Sacks1989}. For a modern review, see
\citet{williams2006gaussian} or \citet{Gramacy2020}. The most popular
acquisition function is {\em expected improvement}
\citep[EI;][]{Jones1998}. EI balances exploration and exploitation by 
suggesting locations with either high variance or low mean, or both.  Greater 
detail is provided in Section \ref{sec:REI}. EI is highly effective and has 
desirable theoretical properties.

\begin{figure}[ht!]
	\centering
	\includegraphics[width=8.75cm,trim=0 45 0 60]{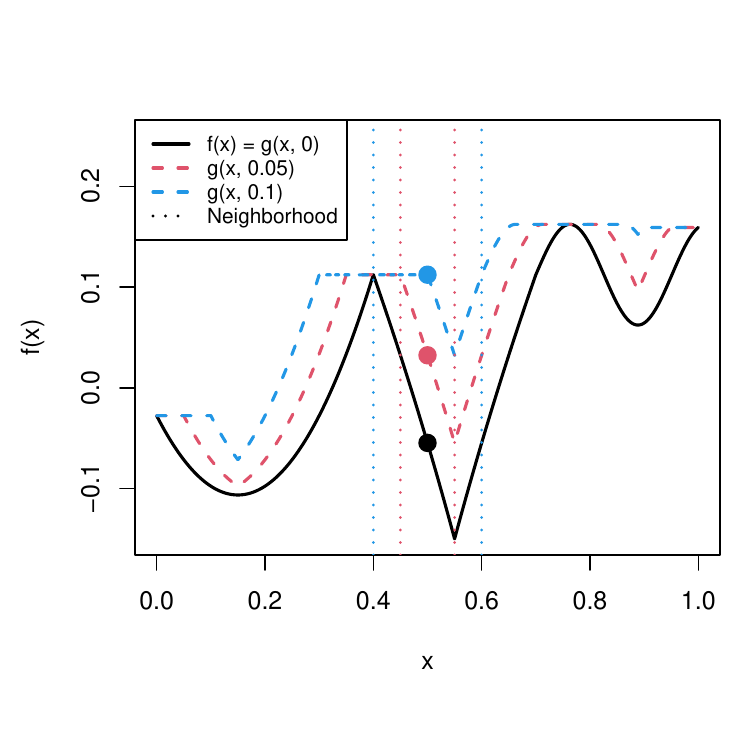}
	\caption{Surface for a 1d multimodal function shown with different 
	adversary levels \blu{and $\alpha$-neighborhoods}.}
	\label{fig:1D-Example}
\end{figure}

Our goal in this paper is to extend BO via GPs and EI to a richer, more
challenging class of optimization problems.  In situations where there are a
multitude of competing, roughly equivalent local solutions to
Eq.~(\ref{eq:bogoal}) a practitioner may naturally express a preference 
\blu{for} ones which enjoy a wider {\em domain of attraction}; i.e., those whose
troughs are larger. \blu{Such preferences for {\em robust} solutions are
expressed across diverse disciplines, e.g., aerospace engineering
\citep{Li2002}, electrical engineering \citep{Kolvenbach2018}, economics
\citep{Goldfarb2003}. Later we demonstrate a real world autonomous warehousing
problem \citep{Kaelbling2017} that benefits from robust design.} For \blu{a
simple} example, consider $f(x)$ defined in Eq.~(\ref{eq:ryan}), provided
later in Section \ref{sec:empcomp}, as characterized by the black line in
Figure \ref{fig:1D-Example}. It has three local minima.  The one around
$x=0.9$ is significantly higher than the other two; \blu{the} one around 0.55
\blu{is quite narrow but has a} lower objective value compared to the
\blu{local minimum} around 0.15. \blu{However, that solution is more robust
because it has it has a wider area of low-values nearby.
}
Although BO via GPs/EI would likely explore both 
domains of attraction to a certain extent {\em eventually}, it will in the near 
term (i.e., for smaller simulation budgets) focus on the deeper/narrower one, 
providing lower resolution on the solution than many practitioners prefer.

Finding \blu{or} exploring the {\em robust} global solution space requires a
modification to both the problem and the BO strategy.  Here we borrow a
framework first introduced in the math programming literature on robust
optimization \citep[e.g.,][]{Menickelly2020}: an {\em adversary}. Adversarial
reasoning is popular in reinforcement learning \citep{Huang2011}.  To our
knowledge, the mathematical programming notion of an adversary has never been
deployed for BO, where it is perhaps best intuited as a penalty on
``sharp'' local minima.   Specifically, let $x^\alpha$ denote the
$\alpha$-{\em neighborhood} of an input $x$.  There are many ways to make this
precise depending on context. If $x$ is one-dimensional, then $x^\alpha =
[x-\alpha, x+\alpha] \equiv [x \pm \alpha]$ is sensible.  In higher dimension,
one can generalize to an $\alpha$-ball or hyper-rectangle. More specifics will
come later. Relative to that $\alpha$-neighborhood, an adversary $g(x,\alpha)$
and robust minimum $x^r$ may be defined as follows:
\begin{equation}
\mbox{Let} \quad g(x, \alpha) =
\max_{x \in x^\alpha} \ f(x) \quad \quad \mbox{so that} \quad \quad 
x^r = \argmin_{x \in [0,1]^d} \ g(x, \alpha). \label{eq:robust}
\end{equation} 
Figure \ref{fig:1D-Example} provides some examples. 
Observe that the larger $\alpha$ is, the more ``flattened'' 
$g(x, \alpha)$ is compared to $f(x)$. In particular, note how the penalization 
is more severe in the sharper minimum compared to the shallow trough which is
only slightly higher than the original, unpenalized function.

The conventional, local, approach to finding the robust solution $x^r$ is best
understood as an embellishment of schemes inspired by Newton's method
\citep[e.g.,][]{Cheney1959}. First, extract derivative (and adversarial)
information nearby through finite differencing and evaluation at the boundary
of $x^\alpha$; then take small steps that descend the (adversarial) surface.
Such schemes offer tidy local convergence guarantees but are profligate with
evaluations. \blu{\cite{Leyffer2020} provide a survey of many of such methods,
e.g., AIMSS \citep{Bisschop1994}, JuMPeR \citep{Dunning2017}, ROC
\citep{Bertsimas2019, Vayanos2022}, ROME \citep{Goh2009}, ROPI
\citep{Goerigk14}, SIPAMPL \citep{Vaz2006}, and YAMIP \citep{Lofberg2008}.}
When $f$ is expensive, \blu{these approaches are} infeasible. We believe that
idea of building an adversary can be ported to the BO framework, making better
use of limited simulation resources toward global optimization while still
favoring wider domains of attraction.

The crux of our idea is as follows.   A GP $\hat{f}_n$, that would typically be
fit to a collection of $n$ evaluations of $f$ in a BO framework, can be used to
define the adversarial realization of those same values following the fitted
analog of $g$: $\hat{g}(x_i, \alpha) = \max_{x \in x_i^\alpha} \hat{f}(x)$. A
second GP\blu{,} $\hat{f}^\alpha_n$\blu{,} can be fit to those $\hat{g}(x_i,
\alpha)$-values, $i=1,\dots,N$.  We call $\hat{f}^\alpha_n$ the {\em adversarial
surrogate}. Then, \blu{one} can use $\hat{f}^\alpha_n$ as you would an ordinary 
surrogate, $\hat{f}_n$, with EI guiding acquisitions.  We call this 
{\em robust expected improvement (REI)}.  There are, of course, myriad details 
and variations -- simplifications and embellishments -- that we are glossing 
over here, and that we shall be more precise about in due course.  The most 
interesting of these may be how we suggest dealing with a practitioner's 
natural reluctance to commit to a particular choice of $\alpha$, {\em a priori}.

Before discussing details, it is worth remarking that the term ``robust'' has
many definitions across the statistical and optimization literature(s). Our
use of this term here is more similar to some than others. We do not mean
robust to outliers \citep{Martinez-Cantin2009} as may arise when noisy,
leptokurtic simulators $f$ are involved \citep{Beland2017} -- though we shall
have some thoughts on this setup later.  \blu{Nor do we mean robust to
uncontrollable input variables, as in {\em robust parameter design}
\citep{taguchi1986introduction}, although again there are similarities.} Some
refer to robustness as the choice of random initialization of a local
optimizer \citep{Taddy2009}. However our emphasis is global. Our definition in
Eq.~(\ref{eq:robust}), and its BO implementation, bears some similarity to
\citet{Oliveira2019} and to so-called ``unscented BO'' \citep{Nogueira2016},
where robustness over noisy or imprecisely specified {\em inputs} is
considered.  However, our adversary entertains a worst-case scenario
\citep[e.g.,][]{Bogunovic2018}, rather than the stochastic/expected-case.
\citet{Marzat2016} also consider worst-case robustness, but focus on so-called
``minimax'' problems where some dimensions are perfectly controlled ($\alpha =
0$) and others are completely uncontrolled ($\alpha = 1$). Our definition of
robustness -- $x^r$ in Eq.~(\ref{eq:robust})
-- emphasizes a worst-case within the $\alpha$\blu{-neighborhood,
   $x^{\alpha}$}. \blu{When data are scaled, $\alpha$ can be thought of as a
   percentage; you want to find the best, worst-case scenario when you can
   control your inputs within $\alpha \times 100\%$.} Nonetheless, these other
   methods and their test cases make for interesting empirical comparisons, as
   we demonstrate. 

The rest of this paper continues as follows. Section \ref{sec:review} reviews
the bare essentials required to understand our idea. Section \ref{sec:methods}
introduces our proposed robust BO setup, via adversaries, surrogate modeling
and REI, and variations. Section \ref{sec:empirical} showcases empirical
results via comparison to \blu{REI} and to similar methodology from
the recent BO literature.  Section \ref{sec:rob} then ports that benchmarking
framework to a \blu{real world} robot pushing problem. Finally, Section
\ref{sec:conc} concludes with a brief discussion.

\section{Review} 
\label{sec:review}

Here we review the basic elements of BO: GPs and EI.

\subsection {Gaussian process regression}
\label{sec:gp} 

A GP may be used to model smooth relationships between inputs and outputs for
expensive black-box simulations as follows. Let $x_i \in [0, 1]^d$ represent
$d$-dimensional inputs and $y_i = f(x_i)$ be outputs, $i=1,\dots, n$.  We
presume $y_i$ are \blu{deterministic realizations of $f$} at $x_i$, though the
GP/BO framework may easily be extended to noisy outputs \citep[see,
e.g.,][Section 5.2.2 and 7.2.4]{Gramacy2020}.  Collect inputs into $X_n$, an
$n \times d$ matrix with rows $x_i^\top$, and outputs into column $n$-vector
$Y_n$. These are the training data\blu{,} $D_n = (X_n, Y_n)$.  Then, model
$Y_n \sim \mathcal{N}_n(\mu(X_n), \Sigma(X_n, X_n))$, i.e., presume outputs
follow an $n$-dimensional multivariate normal distribution (MVN).  \blu{Since
our outputs $y_i$ are deterministic, the Gaussian distribution is not on the
responses -- even though it is notated that way for compactness -- but actually on
on the latent (random) field of function realizations, $f$.}

Often, $\mu(X_n) = 0$ is sufficient for coded inputs and outputs (i.e., after
centering and normalization), moving all of the modeling ``action'' into the
covariance structure $\Sigma(\cdot, \cdot)$, which is defined through the choice
of a positive-definite, distance-based {\em kernel}.  Here, we prefer a
squared exponential kernel, but others such as the Mat\`{e}rn \citep{Stein1999,
Abrahamsen1997} are common. This choice is not material to our presentation;
both specify that the function $f$, via $x_i$ and $y_i$, vary smoothly as a
function of inverse distance in the input space. Details can be found in
\citet[][Section 5.3.3]{Gramacy2020}. Specifically, we fill the covariance
structure $\Sigma_{ij} \equiv \Sigma(X_n, X_n)_{ij}$ via
\begin{equation}
\label{eq:cov}
\Sigma_{ij} = \tau^2 \left(\textrm{exp}\left(-\frac{||x_i - x_j||^2}
	{\theta}\right) + \epsilon\mathbb{I}_{\{i=j\}}\right).
\end{equation}

The structure in Eq.~(\ref{eq:cov}) is hyperparameterized by $\tau^2$ and
$\theta$. Let $\hat{\tau}^2$ and $\hat{\theta}$ be estimates for the
hyperparameters estimated through the MVN log likelihood. Scale $\tau^2$
captures vertical spread between peaks and valleys \blu{of} $f$. Lengthscale 
$\theta$ captures how quickly the function changes direction. Larger $\theta$ 
means the correlation decays less quickly leading to flatter functions. Observe 
that $\epsilon$-jitter \citep{neal1998regression} is added to the diagonal to
ensure numerical stability when decomposing $\Sigma$.

Working with MVNs lends a degree of analytic tractability to many statistical
operations, in particular for conditioning \citep[e.g.,][]{Kalpic2011}  as
required for prediction. Let $\mathcal{X}$, an $n_p \times d$ matrix, store
inputs for ``testing.'' Then, the conditional distribution for
$Y(\mathcal{X})$ given $D_n = (X_n, Y_n)$ is also MVN and has a convenient
closed form:
\begin{align}
	\label{eq:GPpreds}
	&Y(\mathcal{X}) \mid  D_n \sim \mathcal{N}_{n_p}(\mu_n(\mathcal{X}), 
	\Sigma_n(\mathcal{X}))\\
	\mbox{where } \quad &\mu_n(\mathcal{X}) = \Sigma(\mathcal{X}, 
	X_n)\Sigma^{-1}(X_n, X_n)Y_n\nonumber \\
	\mbox{and } \quad &\Sigma_n(\mathcal{X}) = \Sigma(\mathcal{X}, \mathcal{X}) 
	- \Sigma(\mathcal{X}, X_n)\Sigma^{-1}(X_n, X_n)\Sigma(X_n, 
	\mathcal{X}), \nonumber
\end{align}
where $\Sigma(\cdot, \cdot)$ extends $\Sigma_{ij} \equiv \Sigma(x_i, x_j)$ in
Eq.~\eqref{eq:cov} to the rows of $\mathcal{X}$ and between $\mathcal{X}$ and 
$X_n$. The diagonal of the $n_p \times n_p$ matrix $\Sigma_n(\mathcal{X})$ 
provides pointwise predictive variances which may be denoted as
$\sigma_n^2(\mathcal{X})$. Later, we shall use $\text{GP}_{\hat{\theta}}(D_n)$
to indicate a GP surrogate $\hat{f}_n$ \blu{fitted} to data $D_n = (X_n, Y_n)$
emitting predictive equations $(\mu_n(\cdot), \sigma_n^2(\cdot))$ as in
Eq.~(\ref{eq:GPpreds}), conditioned on estimated hyperparameters
$\hat{\theta}_n$.  Note that we are streamlining the notation here
somewhat and subsuming $\hat{\tau}^2$ into $\hat{\theta}$. For more
information on GPs, see \citet[Chapter 5]{Gramacy2020} or \cite{Santner2018}.

\subsection {Expected improvement}
\label{sec:ei} 

Bayesian Optimization (BO) seeks a global minimum (\ref{eq:bogoal}) under a
limited experimental budget of $N$ runs. The idea is to proceed sequentially,
$n = n_0, \dots, N$, and in each iteration $n$ make a {\em greedy} selection
of the next, $n+1^\mathrm{st}$ run, $x_{n+1}$, based on solving an acquisition
function \blu{tied to $\hat{f}_n$}.  The initial $n_0$-sized design could be
space-filling, for example with a Latin hypercube sample (LHS) or maximin
design \citep[Chapter 17]{Dean2015} or, as some have argued
\citep{zhang2018distance}, purely at random.  GPs have emerged as the
canonical \blu{surrogate for BO}.  Although there are many acquisition
functions in the literature tailored to BO via GPs, expected improvement (EI)
is perhaps the most popular. EI may be described as follows.

Let $f^{\min}_n = \min(y_1, \dots, y_{n})$ denote the best ``best observed
value'' (BOV) found so far, after the first $n$ acquisitions ($n_0$ of which
are space-filling/random).  Then, define the {\em improvement} at input
location $x$ as $I(x) = \max\{0, f^{\min}_n - Y(x)\}$.  $I(x)$ is a random
variable, inheriting its distribution from $Y(x) \mid D_n$.  If $Y(x)$ is
Gaussian, as it is under $\text{GP}_{\hat{\theta}}(D_n)$ via
Eq.~(\ref{eq:GPpreds}), then the expectation of $I(x)$ has a closed form:
\begin{equation}
\label{eq:ei}
\mathrm{EI}_n(x) = \mathbb{E}\{I(x) \vert D_n\} = (f^{\min}_n - 
\mu_{n}(x))\Phi\left(\frac{f_{n}^{\min} 
- \mu_{n}(x)}{\sigma_{n}(x)}\right) + 
\sigma_{n}(x)\phi\left(\frac{f_n^{\min} - 
\mu_{n}(x)}{\sigma_{n}(x)}\right),
\end{equation}
where $\Phi$ and $\phi$ are the standard Gaussian CDF and PDF, respectively.
Notice how EI is a weighted combination of mean $\mu_n(x)$ and uncertainty
$\sigma_n(x)$, trading off ``exploitation and exploration.'' The first term of
the sum is high when $\hat{f}(x)$ is much lower than $f_n^{\min}$,
while the second term is high when the GP has high uncertainty at $x$. 
\begin{figure}[ht!]
	\includegraphics[width=8cm,trim=0 25 0 50]{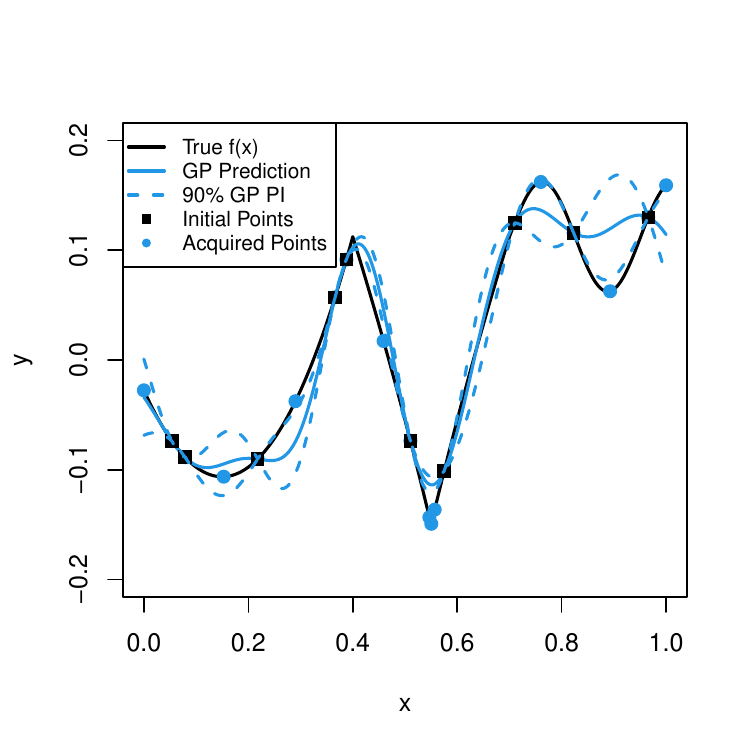}
	\includegraphics[width=8cm,trim=0 25 0 50]{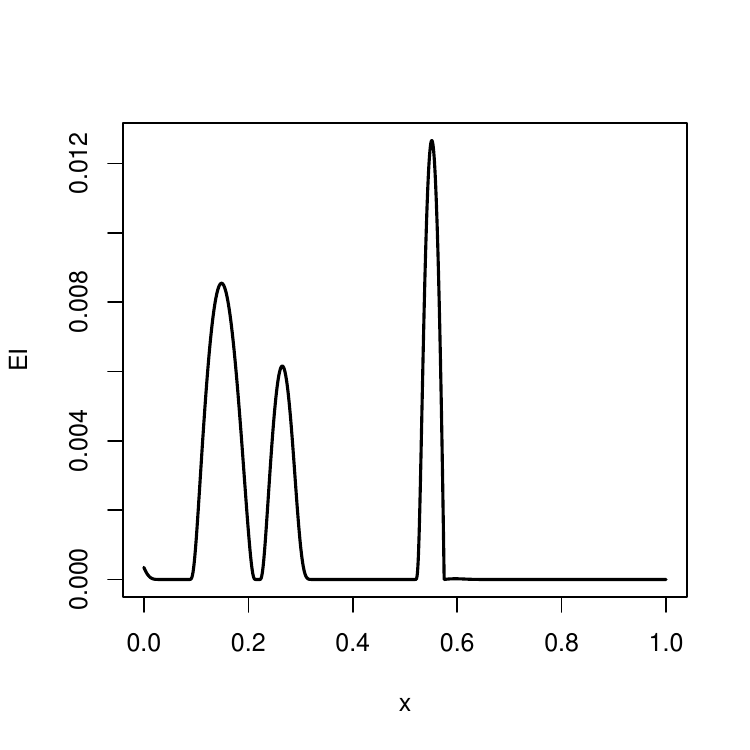}
	\caption{Left: EI-based (EGO) acquisition with $n_0 = 10$ initial points as
	black squares and 10 acquisitions as blue circles using EGO on 
	the 1D example function from Eq.~(\ref{eq:ryan}) with initial GP fit in 
	blue. Right: EI using the initial design at different locations of $x$.}
	\label{fig:EI-Example}  
\end{figure}

The {\em right} panel of Figure \ref{fig:EI-Example} shows an EI surface
implied by the GP surrogate (blue) shown in the {\em left} panel.  This fit is
derived from $n=10$ training data points (black squares) sampled from our
$f(x)$ from Figure \ref{fig:1D-Example}.  We will discuss the blue circles
momentarily.  Observe how EI is high in the two local troughs of minima, near
$x = 0.15$ and $x = 0.55$, but away from the training data. The total volume
(area under the curve) of EI is larger around the left, shallower trough, but 
the maximal EI location is in the right, spiky trough. This ($x$ near 0.55) is
where EI recommends choosing the next, $n+1^\mathrm{st}$ acquisition.

Operationalizing that process, i.e., numerically solving for the next
acquisition involves its own, ``inner'' optimization: $x_{n+1} = \argmax_{x
\in [0,1]^d} \; \mathrm{EI}_n(x)$. This can be challenging because EI is
multi-modal.  Observe that while $f$ only has two local minima, EI as shown
for $n=10$ in Figure \ref{fig:EI-Example}, has three local maxima.  So in some
sense the inner optimization is harder than the ``outer'' one.  As $n$ grows,
the number of local EI maxima can grow.   A numerical optimizer such as BFGS
\citep{Byrd1995} is easily stuck, necessitating a multi-start scheme
\citep{Burden1989}. Another common approach is to use a discrete,
space-filling set of candidates, \blu{turning} a continuous search for $x \in
[0,1]^d$ into a discrete one for $x \in
\mathcal{X}_{\mathrm{cand}}$.  Hybrid and smart-candidate schemes are also
popular \citep{Scott2011, Gramacy2022}.  In general, we can afford a
comprehensive effort toward solving the ``inner'' optimization because the
objective, derived from $\text{GP}_{\hat{\theta}}(D_n)$, is cheap -- especially
compared to the black-box $f(\cdot)$. Repeated application of EI toward
selecting $x_{n+1}$ is known as the {\em efficient global optimization} 
algorithm \citep[EGO,][]{Jones1998}.

The blue circles in Figure \ref{fig:EI-Example} indicate how nine further
acquisitions (ten total) play out, i.e., $10 = n_0, \dots, n, \dots, N=20$ in
EGO.  Notice that the resulting training data set $D_N$, combining both black 
and blue points concentrates acquisitions in the ``tip'' of the spiky, right 
trough.  The rest of the space, including the shallower left trough, is more 
uniformly (and sparsely) sampled.  Our goal is to concentrate more acquisition 
effort in the shallower trough.

\section{Proposed methodology}
\label{sec:methods}

Here we extend the EGO algorithm by incorporating adversarial thinking. The
goal is to find $x^r$ from Eq.~(\ref{eq:robust}) for some $\alpha$.  We begin
by presuming a fixed, known $\alpha$ selected by a practitioner.

\subsection {The adversarial surrogate}
\label{sec:adv}

If $\hat{f}_n(x)$ is a surrogate for $f(x)$, then one may analogously notate
$\hat{f}_n^\alpha(x)$ as a surrogate for $g(x, \alpha)$, an adversarial
surrogate.  We envision several ways in which $\hat{f}_n^\alpha(x)$ could be
defined in terms of $\hat{f}_n(x)$, but not many which are tractable to work
with analytically or numerically.  For example, suppose $Y^\alpha(x) =
\max_{x' \in x^\alpha} Y(x')$, where $Y(x') \sim \mathcal{N}(\mu_n(x'),
\sigma^2_n(x'))$, a random variable whose distribution follows the spirit 
of Eq.~(\ref{eq:robust}), but uses the predictive equations of 
Eq.~(\ref{eq:GPpreds}). The distribution of $Y^\alpha(x)$ could be a version of 
$\hat{f}_n^\alpha(x)$, at least notionally.  However a closed form remains 
elusive.

Instead, it is rather easier to define $\hat{f}_n^\alpha(x)$ as an ordinary
surrogate trained on data derived through adversarial reasoning on the
original surrogate $\hat{f}_n(x)$.  \blu{Let $Y_n^\alpha$ denote these {\em
adversarial responses}, where each $y_i^\alpha$, for $i=1,\dots,n$, follows}
\begin{equation}
	\label{eq:advy}
	y_i^\alpha = \max_{x \in x^\alpha_i} \; \mu_n(x) \quad \mbox{where 
	$x_i^\alpha$ is the $\alpha$-neighborhood of the $i^\mathrm{th}$ entry of 
	$X_n$, as usual.} 
\end{equation}
There are many sensible choices for finding $y_i^\alpha$ numerically.
Newton-based optimizers, e.g., BFGS, could leverage closed form derivatives
for $\mu_n(x)$ finite differencing. A simpler option that works well is to
instead take $y_i^\alpha = \max_{x \in x_i^{\mathcal{B}_\alpha(x)}}
\mu_{n}(x)$, where $\mathcal{B}_{\alpha}(x)$ is the discrete set of points on
the corners of a box with sides of length $\alpha$ emanating from $x$. Details
for our own implementation are deferred to Section \ref{sec:implement}.

Given $Y_n^\alpha$, an adversarial surrogate may be built by
modeling {\em adversarial data} $D_n^\alpha = (X_n, Y_n^\alpha)$, \blu{i.e.,
$Y_n^\alpha$ paired} with their original inputs, as a GP. Let
$\hat{\theta}_\alpha$ denote hyperparameter estimates for
$\hat{f}_n^\alpha(x)$. Fill in the covariance matrix following
Eq.~(\ref{eq:cov}), $\Sigma^\alpha(\cdot, \cdot)$. Finally, define the
{\em adversarial surrogate} as
\begin{equation}
	\label{eq:advsurr}
	\hat{f}_n^\alpha(x) \equiv \text{GP}_{\hat{\theta}_\alpha}(D_n^\alpha) \rightarrow (\mu_n^\alpha(\cdot), \sigma_n^{2\alpha}(\cdot))
\end{equation}
via novel hyperparameter estimates $\hat{\theta}_\alpha$ using $Y_n^\alpha$
rather than $Y_n$ with predictive equations akin to (\ref{eq:GPpreds}). Since 
the $Y_n^\alpha$ are the original surrogate's ($\hat{f}_n$) estimate of
adversarial response values according $f_\alpha$, $\hat{f}_n^\alpha$ may serve
as a surrogate for $g(x,\alpha)$ from the left half of Eq.~(\ref{eq:robust}).
\begin{figure}[ht!]
	\centering
	\includegraphics[width=8cm,trim=0 20 0 50]{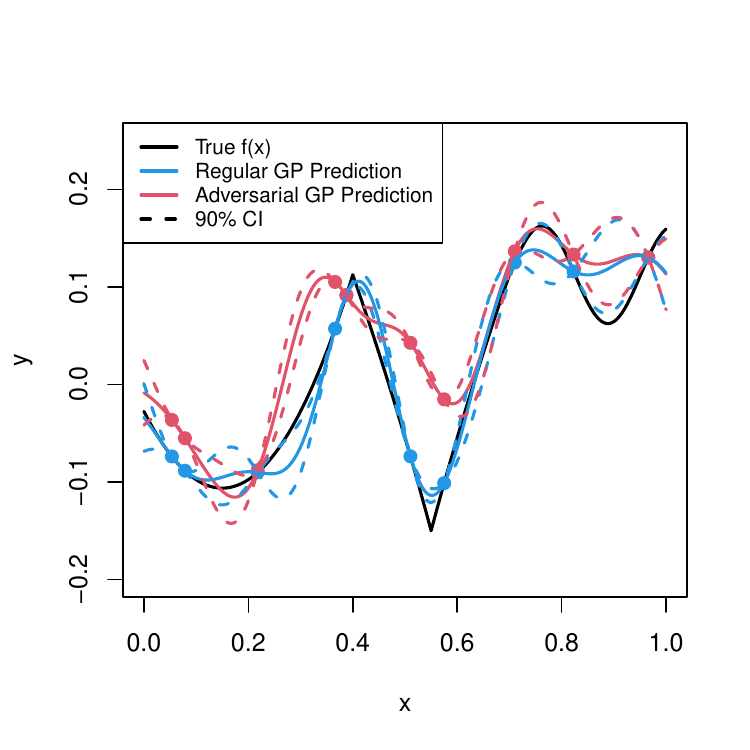}
	\includegraphics[width=8cm,trim=0 20 0 50]{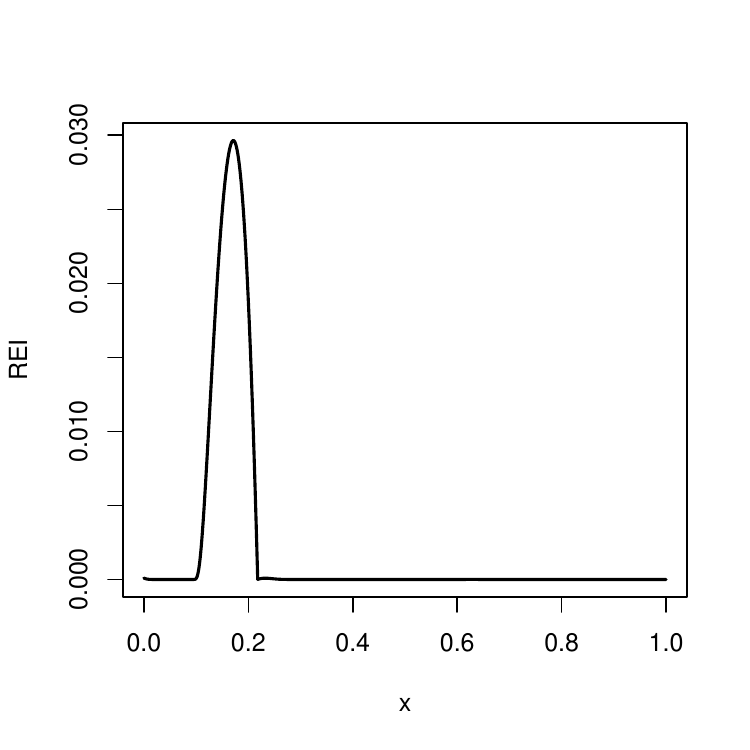}
	\caption{Left: The same black and blue lines from Figure 
	\ref{fig:EI-Example}. In red, $Y_n^\alpha$ are shown as the points with 
	$\hat{f}_n^\alpha(x)$ as the curve. Right: Robust expected improvement for 
	the current design.}
	\label{fig:Adv_Surr}
\end{figure}

As a valid GP, $\hat{f}_n^\alpha$ can be used in any way another surrogate
might be deployed downstream. For example, $\hat{f}_n^\alpha$ may be used to
acquire new runs via EI, but hold that thought for a moment. To illustrate
$\hat{f}_n^\alpha$, the left panel of Figure \ref{fig:Adv_Surr} augments the
analogous panel in Figure \ref{fig:EI-Example} to include a visual of
$\hat{f}_n^\alpha$ via $\mu_n^\alpha$ and error-bars $\mu_n^\alpha \pm
\sigma_n^\alpha$. Notice that the new, red lines, are much higher near the
sharp minimum (near $x=0.5$), but less dramatically elevated for the wider,
dull trough near $x=0.2$.

Also observe in Figure \ref{fig:Adv_Surr} that, whereas for the most part 
$\mu_n(x) \leq \mu^\alpha_n(x)$, this reverses for a small portion of the input 
domain near $x=0.2$. That would never happen when comparing $f(x)$ to $g(x, 
\alpha)$ via Eq.~(\ref{eq:robust}).  This happens for our 
surrogates -- original $\hat{f}_n$ and adversarial $\hat{f}_n^\alpha$ -- for 
two reasons.  One is
   that both are stationary processes because their covariance structure
   (\ref{eq:cov}) uses only relative distances, resulting in a compromise
   surface between sharp and dull.  Although this is
   clearly a mismatch to our data-generating mechanism, we have not found any
   downsides in our empirical work. \blu{Possibly improved
   performance for non-stationary surrogates is entertained in Section
   \ref{sec:conc}.}  The other reason is that the adversarial data
   $Y_n^\alpha$, whether via a stationary or (speculatively) a non-stationary
   surrogate $\hat{f}_n$, provide a low-resolution view of the true adversary
   $g(x, \alpha)$ when $n$ is small.  Consequently, $\hat{f}_n^\alpha$ is a
   crude approximation to $g(x, \alpha)$, but that will improve with
   more acquisitions (larger $n$).  What is most important is the EI surface(s) 
   that the surrogate(s) imply.  These are shown in the right panel of Figure 
   \ref{fig:Adv_Surr}, and discussed next.

\subsection {Robust expected improvement} 
\label{sec:REI} 

{\em Robust expected improvement} (REI), $\mathrm{EI}_n^\alpha(\cdot)$, is the
analog of in Eq.~(\ref{eq:ei}) except using $(\mu_n^\alpha(\cdot),
\sigma_n^{2\alpha}(\cdot))$, towards solving the mathematical program given
in the right half of Eq.~(\ref{eq:robust}). Let $f_n^{\alpha,\min} = 
\min(y_1^\alpha, \dots, y_{n}^\alpha)$ denote the best estimated adversarial 
response (BEAR). Then, let
\begin{equation}
	\label{eq:advei}
	\text{EI}^\alpha_n(x) =  \mathbb{E}\{I(x) \mid D_n^\alpha\} = 
	(f_{n}^{\alpha,\min} - \mu_{n}^\alpha(x))\Phi\!\left(\frac{f_{n}^{\alpha,\min} - \mu_{n}^\alpha(x)}{\sigma_{n}^\alpha(x)}\right) + 
	\sigma_{n}^\alpha(x)\phi\!\left(\frac{f_{n}^{\alpha,\min} - 
	\mu_{n}^\alpha(x)}{\sigma_{n}^\alpha(x)}\right).
\end{equation} 
The right panel of Figure \ref{fig:Adv_Surr} \blu{shows an REI surface arising
from} the same initial setup as Figure \ref{fig:EI-Example}. We see only one
peak, located at the shallower, wider trough compared to three peaks \blu{for
EI} [Figure \ref{fig:EI-Example}]. Generally speaking, REI surfaces have fewer
local maxima compared to EI because $\hat{f}_n^\alpha(x)$ smooths over the
peaked regions. Thus the inner optimization of REI often has fewer local
minima, requiring fewer multi-starts.

REI is summarized succinctly in Eq.~(\ref{eq:advei}) above, but it is
important to appreciate that it is a result of a multi-step process. 
Alg.~\ref{alg:REI_Fixed} provides the details: fit an ordinary surrogate, which 
is used to create adversarial data, in turn defining the adversarial surrogate 
upon which EI is evaluated. The algorithm is specified for a particular 
reference location $x$, used toward solving $x_{n+1} = \argmax_{x \in [0,1]^d} 
\; \text{EI}_n^\alpha(x)$. It may be applied identically for any $x$.  In a 
numerical solver it may be advantageous to cache quantities unchanged in $x$.

\medskip
\begin{algorithm}[H] 		
\DontPrintSemicolon
\textbf{input} $D_{n} = (X_{n}, Y_{n})$, $x$, and $\alpha$\\
$\hat{f}_n(x) = \text{GP}_{\hat{\theta}}(D_n)$ \tcp*{with predictive moments 
	in Eq.~(\ref{eq:GPpreds})}
	\For{i = 1, $\dots$, $n$} {
		$y_i^\alpha$ = ${\argmax_{x' \in x^\alpha_i}} \ \mu_n(x')$ 
		\tcp*{adversarial responses, (\ref{eq:advy})}
	}
	$D_n^\alpha = (X_n, Y_n^\alpha) = (X_n, \{y_i^\alpha\}_{i=1}^n)$\;
	$\hat{f}^\alpha_n(x) = \text{GP}_{\hat{\theta}_\alpha}(D_n^\alpha)$ 
	\tcp*{the adversarial surrogate, (\ref{eq:advsurr})}
\Return $(\text{EI}_n^\alpha(x))$ \tcp*{EI using 
$\hat{f}^\alpha_n(x)$ from the previous line, (\ref{eq:advei})}
\caption{Robust Expected Improvement}
\label{alg:REI_Fixed}
\end{algorithm}
\noindent 
\medskip

With repeated acquisition via EI, over $n=n_0,\dots,N - 1$ being known as EGO,
we dub robust efficient global optimization (REGO) as the repeated
application of REI towards finding a robust minimum. REGO involves a loop over
Alg.~\ref{alg:REI_Fixed}, with updates $D_n
\rightarrow D_{n+1}$ after each acquisition. The final data set, $D_N = (X_N, 
Y_N)$ provides insight into both $f(x)$ and $g(x,\alpha)$ and their minima.  
Whereas EGO would report BOV $f_N^{\min}$, and/or the corresponding element 
$x^\star_{\mathrm{bov}}$ of $X_N$, REGO would report the BEAR 
$f_N^{\alpha,\min}$, the same quantity used to define REI in 
Eq.~(\ref{eq:advei}), and/or input $x^\star_{\mathrm{bear}}$.

\subsubsection*{Post hoc adversarial surrogate}

To quantify the advantages of REI/REGO over EI/EGO in our
empirical work of Sections \ref{sec:empirical}--\ref{sec:rob}, we consider
a {\em post hoc adversarial surrogate}.  This is the surrogate constructed after
running all acquisitions, $n_0 + 1,\dots,N$ via EI/EGO, then at $N$ fitting an
adversarial surrogate Eq.~(\ref{eq:advsurr}), and extracting the BEAR
$f_N^{\alpha, \min}$ rather than the BOV.  In other words, the last step is
faithful to the adversarial goal, whereas active learning aspects ignore it
and proceed as usual.  \blu{While} the BOV from EGO can be a poor approximation to
the robust optimum $f(x^r)$ of Eq.~(\ref{eq:robust}), the BEAR from a post hoc
adversarial surrogate can potentially be better. Comparing two BEAR solutions,
one from REGO and one from a post hoc EGO surrogate, allows us to separately
explore the value of REI acquisitions from post hoc adversarial surrogates,
$\hat{f}_N^\alpha$.

\begin{figure}[ht!]
	\centering
	\includegraphics[width=9.8cm,trim=0 43 0 30]{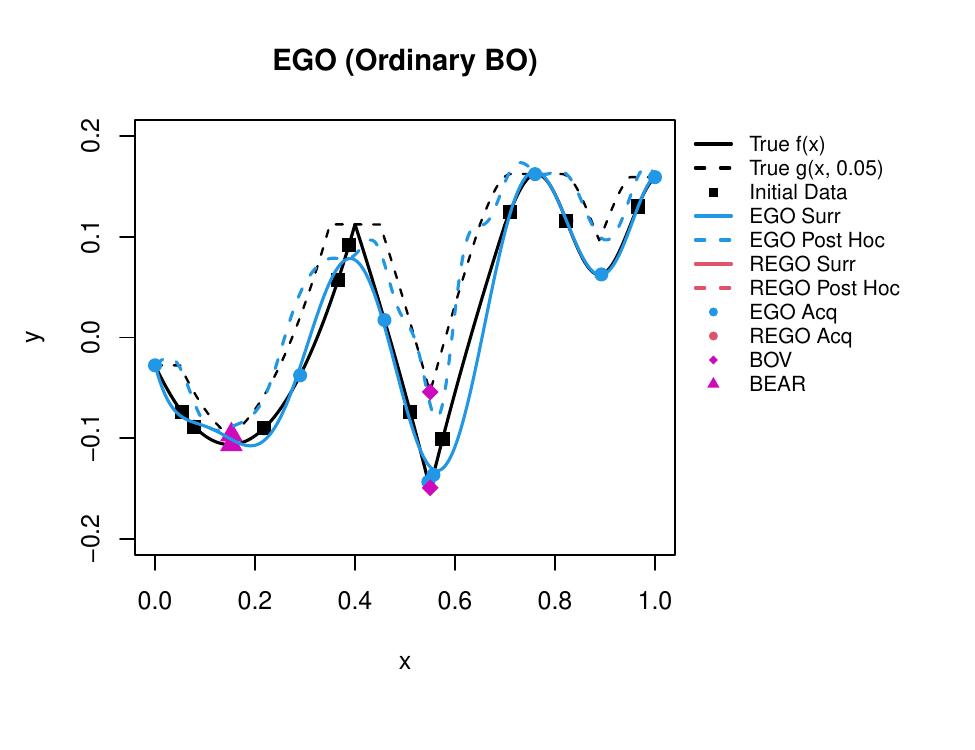}
	\includegraphics[width=7cm,trim=2 22 0 35]{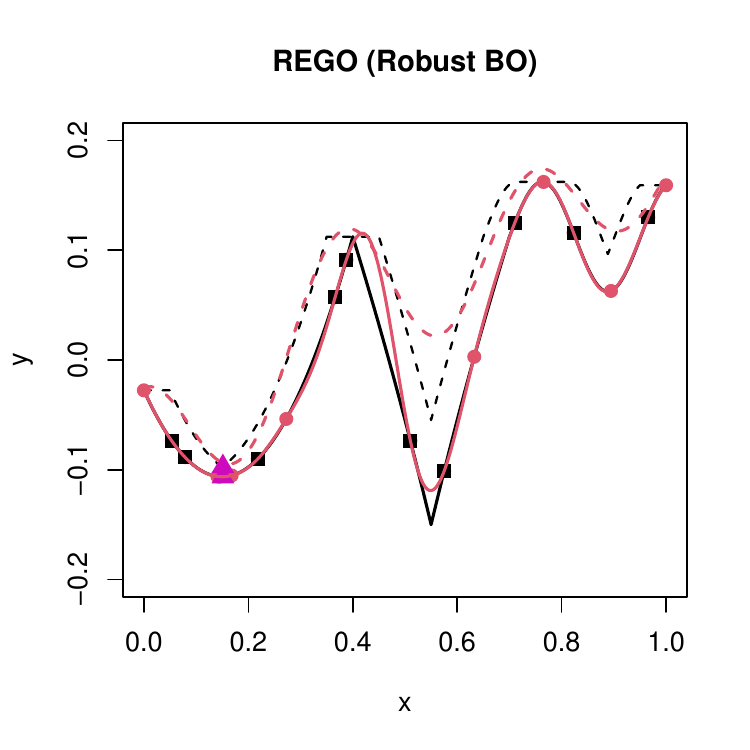}
	\caption{Left: Surrogate and post hoc surrogate fits after EGO acquisitions
	($N=20$ and $n_0 = 10$) on Eq.~(\ref{eq:ryan}).  Right: Same except
	REGO where BOV and BEAR are the same.}
	\label{fig:posthoc}
\end{figure}

To illustrate on our running example, consider the outcome of EGO,
from the left panel of Figure \ref{fig:EI-Example}, recreated in
the Figure \ref{fig:posthoc} (left).  In both, the union of black and
blue points comprise $D_N$.  In Figure \ref{fig:posthoc} the dashed blue
curve provides the post hoc adversarial surrogate, $\hat{f}_N^{\alpha}$, from
these runs, with the BOV and BEAR indicated as magenta diamonds and triangles, 
respectively. Notice that for EGO, the BOV is around the global,
peaked minimum, but the BEAR captures the shallow, robust minimum. The
analogous REGO run, via EI acquisitions (red circles) and adversarial
surrogate (red lines) is shown in the right panel of Figure \ref{fig:posthoc}.
Here BEAR and BOV estimates are identical since REGO does not explore the
peaked minimum.

Looking at both panels of Figure \ref{fig:posthoc}, the BEAR offers a robust
solution for both EGO or REGO. However, EGO puts only one of its acquisitions
in the wide trough compared to four from REGO.  Consequently, REGO's BEAR
offers a more accurate estimate for $x^r$. With only $N=20$ points in 1d, any
sensible acquisition strategy should yield a decent meta-model for $f(x)$ and
$g(x, \alpha)$, so we may have reached the limits of the utility of this
illustrative example. In Sections \ref{sec:empirical}--\ref{sec:rob}
we intend to provide more compelling evidence that REGO is better at targeting
robust minima. Before turning to that empirical study, we introduce two REI
variations that feature in some of those examples.

\subsubsection*{Unknown and vector-valued $\alpha$} 

Until this point we have presumed fixed, scalar $\alpha$, perhaps specified by
a practitioner.  In the case of inputs $x$ coded to the unit cube $[0,1]^d$, a
choice of $\alpha = 0.05$, say, represents a robustness specification of 5\%
in each coordinate direction (totaling 10\% of each dimension), i.e., where
$x^\alpha \equiv [ x_j \pm \alpha : j=1,\dots, d]$.  This intuitive
relationship can be helpful in choosing $\alpha$, however one may naturally
wish to be ``robust'' to this specification. Suppose instead we had an
$\alpha_{\max}$ in mind, where $0 \leq \alpha \leq \alpha_{\max}$. \blu{If REI
is robust to every value $0 \leq \alpha \leq \alpha_{\max}$, then a slight
misspecification in $\alpha_{\max}$ is still robust over most of the same
$\alpha$ values. Whereas a slight misspecification in pinpointing $\alpha$ may
lead to a completely different response.} One option \blu{that allows us be
robust over the range of $0$ to $\alpha_{\max}$} is to simply run REGO with
$\alpha \leftarrow
\alpha_{\max}$, i.e., basing all REI acquisitions on $\alpha_{\max}$ instead. 
Although we do not show this to save space, performance (via BEAR) of 
$\alpha_{\max}$ relative to a nominal $\alpha$ rapidly deteriorates as the gap 
between $\alpha_{\max}$ and $\alpha$ widens.

A better method is to base REI acquisitions on {\em all} $\alpha$-values
between zero and $\alpha_{\max}$ by integrating over them:
$\overline{\mathrm{EI}}^{\alpha_{\max}}_n = \int_0^{\alpha_{\max}}
\text{EI}^\alpha_n(x) \; d F(\alpha)$, say over uniform measure $F$.  We are
not aware of an analytically tractable solution, in part because embedded in
$\text{EI}^\alpha_n(x)$ are complicated processes such as determining
adversarial data, and surrogates fit thereupon. But quadrature is relatively
straightforward, either via Monte Carlo (MC) $\alpha^{(t)} \sim F$, for
$t=1,\dots,T$, over a grid $\{0 = \alpha^{(1)}, \dots,
\alpha^{(T)} = \alpha_{\max}\}$.  Define
\begin{equation}
\overline{\mathrm{EI}}^{\alpha_{\max}}_n \approx \frac{1}{T} \sum_{t=1}^T
\text{EI}^{\alpha^{(t)}}_n(x) 
\quad \mbox{ for the former,} \quad \mbox{or} \quad
\overline{\mathrm{EI}}^{\alpha_{\max}}_n \approx \frac{1}{T} \sum_{t=1}^T
\text{EI}^{\alpha^{(t)}}_n(x) F(\alpha^{(t)}),
	\label{eq:sumei}
\end{equation}
for the latter.   Both may be implemented by looping over
Alg.~\ref{alg:REI_Fixed}  with different $\alpha^{(t)}$ values, and then
averaging to take $x_{n+1} = \argmax_{x\in[0,1]^d}
\overline{\text{EI}}^{\alpha_{\max}}_n$.   Both approximations
improve for larger $T$.

With random $\alpha^{(t)} \sim F$, even a single draw
($T=1$) suffices as an unbiased estimator -- though clearly with high
variance. This has the advantage of a simpler implementation, and faster
execution as no loop over Alg.~\ref{alg:REI_Fixed} is required. In our 
empirical work to follow, we refer to this simpler option as the ``rand'' 
approximation, and the others (MC or grid-based) as ``sum''.  It is remarkable 
how similarly these options perform, relative to one another and to a nominal 
REI with a fixed $\alpha$-value. One could impose $\alpha_{\min} > 0$ 
similarly, though we do not entertain this variation here.

As a somewhat orthogonal consideration, it may be desirable to entertain
different levels of robustness -- different $\alpha$-values $\{\alpha_1, 
\dots, \alpha_d\}$ -- in each of the $d$ coordinate directions. The only change 
this imparts on the description above is that now $x^\alpha = [ x_j \pm 
\alpha_j : j=1,\dots, d]$, a hyperrectangle rather than 
hypercube. Other quantities such as $g(x, \alpha)$ and $\text{EI}_n^\alpha(x)$
remain as defined earlier with the understanding of a vectorized $\alpha =
\{\alpha_1, \dots, \alpha_d\}$ under the hood. Using vectorized
$\alpha_{\max}$ in Eq.~(\ref{eq:sumei}) requires draws from a multivariate
$F$, which may still be uniform, or a higher dimensional grid of
$\alpha^{(t)}$, recognizing that we are now approximating a higher dimensional
integral.  \blu{We note that allowing an $\alpha_j$ to span the entire range
of the $j^\mathrm{th}$ coordinate, while other $\alpha_k$ are zero, mimicks a
setup similar to robust parameter design \citep{taguchi1986introduction}.
However, our BO context is quite different from classical response surface
methods \citep[RSMs, e.g.,][]{myers2016response}.  In particular, RSMs involve
second-order, local models rather than global surrogates.}

\section{Implementation and benchmarking} 
\label{sec:empirical}

Here we provide a description of our implementation, variations,
methods of our main competitors, evaluation metrics and ultimately provide a 
suite of empirical comparisons.  

\subsection{Implementation details}
\label{sec:implement}

Our main methods (REI/REGO, variations and special cases) are coded in
\textsf{R} \citep{R2021}.  These codes may be found, alongside those of our
comparators and  all empirical work in this paper, in our public Git
repository:
\url{https://bitbucket.org/gramacylab/ropt/}.
GP surrogate modeling for our new methods -- our competitors may leverage
different subroutines/libraries -- is provided by the \texttt{laGP} package
\citep{Gramacy2016}, on CRAN.  Those subroutines, which are primarily
implemented in {\sf C}, leverage squared exponential covariance structure
(\ref{eq:cov}), and provide analytic $\hat{\tau}^2$ \blu{using maximum
likelihood estimation} conditional on lengthscale $\theta$.  In our \blu{main
suite of} experiments, $\theta$ is fixed to appropriate values in order to
control MC variation that otherwise arises in repeated MLE-updating of
$\hat{\theta}$, especially in small-$n$ cases.  \blu{In Appendix
\ref{app:esttheta} we demonstrate how these results change when $\hat{\theta}$
is estimated.} More details are provided as we introduce our test cases, with
further discussion in Section \ref{sec:conc}. Throughout we use $\epsilon =
10^{-8}$ in Eq.~(\ref{eq:cov}), as appropriate for deterministic blackbox
objective function evaluations. For ordinary EI calculations we use add-on
code provided by \citet[][Chapter 7.2.2]{Gramacy2020}.  All empirical work was
conducted on an 8-core hyperthreaded Intel i9-9900K CPU at 3.60 GHz with Intel
MKL linear algebra subroutines.

Section \ref{sec:adv} introduced several possibilities for solving
$y_i^\alpha$, whose definition is provided in Eq.~(\ref{eq:advy}).  Although
numerical optimization, e.g. BFGS, is a gold standard, in most cases we found
this to be overkill, resulting in high runtimes for all $(N - N_0)$
acquisitions with no real improvements in accuracy over the following, far
simpler alternative.  We prefer quickly optimizing over the discrete set
$x^{\mathcal{B}_\alpha(x)}$ comprising of a box extending $\alpha$-units out
in each coordinate direction from $x$, or its vectorized analog as described
in Section \ref{sec:REI}.  This ``cornering'' alternative occasionally yields
$y_i^\alpha < y_i$, which is undesirable, but this is easily mitigated by
augmenting the box $x^{\mathcal{B}_\alpha(x)}$ to contain a small number of
intermediate, grid locations in each coordinate direction.  Using an odd
number of such intermediate points ensures $y_i^\alpha \geq y_i$. \blu{For 1d
problems, we use three intermediate points and in higher dimension we increase
this to five.} We find that a $d$-dimensional grid formed from the outer
product for each coordinate (i.e., a $2^5=32$-point-grid for $d=2$)
facilitates a nice compromise between computational thrift, and accuracy of
adversarial $y_i^\alpha$-values compared to the cumbersome BFGS-based
alternative.

In our test problems, which are introduced momentarily and utilize inputs
coded to $[0,1]^d$, we compare each of the three variations of REGO described
in Section \ref{sec:methods}. These comprise: (1) REI with known $\alpha$
(Alg.~\ref{alg:REI_Fixed}); (2) novel $\alpha\sim
\mathrm{Unif}(0,\alpha_{\max})$ at acquisition; and (3) averaging over a
sequence of $\alpha \in [0,\alpha_{\max}]$ (\ref{eq:sumei}). For all examples,
we set $\alpha_{\max} = 0.2$ and average over five equally spaced values. In
figures, these methods are denoted as ``known'', ``rand'' and ``sum'',
respectively. After each acquisition, we use the post hoc adversarial
surrogate to find the BEAR operating conditions: $x^*_{\mathrm{bear}}$ and
$f_{N}^{\alpha,\mathrm{min}}$ in order to track progress.

As representatives from the standard BO literature, we compare against the
following ``straw-men'': EGO \citep{Jones1998}, where acquisitions are based on
EI (\ref{eq:ei}; and ``EY'' \citep{Gramacy2020} which works similarly to EGO,
except acquisitions are selected minimizing $\mathbb{E}[Y(x) \mid D_N]$:
$x^{\text{EY}}_{\text{new}} = \argmin_{x \in [0, 1]^d} \mu_n(x)$. In other
words, EY acquires the point that the surrogate predicts has the lowest mean.
Since it does not incorporate uncertainty ($\sigma_n(x)$), repeated
EY acquisition often stagnates in one region -- usually a local minima
-- rather than exploring new areas like EI does. 
In our 
figures, these comparators are indicated, as follows: ``ego'', 
and ``ey'' respectively for EGO and EY with progress measured by BOV: 
$x^*_{\mathrm{bov}}$ and $f_{N}^{\mathrm{min}}$. Finally, we consider the post 
hoc adversary with progress measured by BEAR for EGO (``egoph'') and uniform 
random sampling (``unif''). In the figures, REI methods are solid curves with 
robust competitors dashed and regular BO dotted.

Our final comparator is {\tt StableOPT} from \cite{Bogunovic2018}. For
completeness, we offer the following by way of a high-level overview; details
are left to their paper. \citeauthor{Bogunovic2018}~assume a fixed, known
$\alpha$, although we see no reason why our extensions for unknown $\alpha$
could not be adapted to their method as well. Their algorithm relies on
confidence bounds to narrow in on $x^r$.  Let $\mathrm{ucb}_n(x) = \mu_n(x) +
2\sigma_n(x)$ denote the upper 95\% confidence bound at $x$ for a fitted
surrogate $\hat{f}_n$, and similarly $\mathrm{lcb}_n(x) = \mu_n(x) -
2\sigma_n(x)$ for the analagous lower bound. Then we may translate their
algorithm into our notation, shown in Alg.~\ref{alg:stableopt}, furnishing the
$n^{\mathrm{th}}$ acquisition. Similar to REGO, this may then be wrapped in a
loop for multiple acquisitions. We could not find any public software for {\tt
StableOPT}, but it was relatively easy to implement in {\sf R}; see our public
Git repo.

\bigskip
\begin{algorithm}[H] 		
	\DontPrintSemicolon
	\textbf{input} $D_{n - 1} = (X_{n - 1}, Y_{n - 1})$ and $\alpha$\\
	$\hat{f}_{n - 1}(x) = \mathrm{GP}_{\hat{\theta}}(D_{n - 1})$ \tcp*{with 
	predictive moments in Eq.~(\ref{eq:GPpreds})}
	$\tilde{x}_n = \argmin_{x \in [0, 1]^d} \max_{a \in [-\alpha, \alpha]} 
	\mathrm{lcb}_{n - 1}(x + a)$ \tcp*{where we think $x^r$ is}
	$a_n = \argmax_{a \in [-\alpha, \alpha]} \mathrm{ucb}_{n - 1}(\tilde{x}_n + 
	a)$ \tcp*{worst point within $\tilde{x}_n^\alpha$}
	\Return $(\tilde{x}_n + a_n, f(\tilde{x}_n + a_n)))$ \tcp*{sample from $f$ 
	and return}
	\caption{{\tt StableOPT}}
	\label{alg:stableopt}
\end{algorithm}
\noindent 
\bigskip

Rather than acquiring new runs nearby likely $x^r$, {\tt StableOPT} samples
the worst point within $\tilde{x}_n^\alpha$. Consequently, its final $X_N$
does not contain any points thought to be $x^r$. \citeauthor{Bogunovic2018}~recommend selecting $x^*_{\mathrm{bear}}$ from all $\tilde{x}_n$ rather than
the actual points sampled, all $\tilde{x}_n + a_n$, using notation introduced
in Alg.~\ref{alg:stableopt}. We generally think it is a mistake to report an
answer at an untried input location.  So in our experiments we
calculate $x^*_{\mathrm{bear}}$ for {\tt StableOPT} as derived from a final,
post hoc adversary calculation [Section \ref{sec:REI}]. In figures, this
comparator is denoted as ``stable''.

\blu{It is worth noting that there are many additional {\em potential} comparators
from the math programming community. Some where mentioned in Section
\ref{sec:intro}. Others include ones developed by \cite{Menickelly2018,
Bertsimas2010, Conn2012, Bertsimas2010b}.  All involve a test case that
is some variation our ``Bertsimas'' example in Eq.~(\ref{eq:bert}), coming
momentarily.  However, it is clear at a glance that these papers' results in
this case are not competitive against our BEAR benchmark.  This is because
those methods were developed with different goals in mind.  For example, some
are inherently local while REI/BO are global. Others require hundreds or
thousands of function evaluations to find the true robust minimum. Our budgets
shown here are in the dozens.  In some cases extensive evaluation makes their
methods more precise, but also more profligate with runs. When the goal is to
optimize an expensive blackbox function, more evaluation is often impossible
and we have to make the most of every acquisition.  Finally, none of the
robust methods that we found from the math programming literature came with an
{\sf R} implementation.  Consequently, we did not include them in our
empirical study.}

The general flow of our benchmarking exercises coming next
[Section \ref{sec:empcomp}] is as follows. Each optimization, say in
input dimension $d$, is seeded with a novel LHS of size $n_0 = 5 + 5d$ that is
shared for each comparator.  Acquisitions, $n=n_0+1,\dots, N$, separate for
each comparator up to a total budget $N$ (different for each problem), are
accumulated and progress is tracked along the way.  Then this is repeated, for
a total of 1,000 MC trials. To simplify notation, let
$x^*_{\mathrm{b, n}}$ be either $x^*_{\mathrm{bear}}$ or $x^*_{\mathrm{bov}}$,
depending on if using BEAR or BOV, after the $n^{\mathrm{th}}$ acquisition. We
utilize the following two metrics to compare BEAR or BOV across methods using
the true adversary:
\begin{equation}
	\label{eq:MCmetrics}
	r(x^*_{\mathrm{b, n}}) = g(x^*_{\mathrm{b, n}}, \alpha) - g(x^r, \alpha)
	\quad \quad \quad \quad
	d(x^*_{\mathrm{b, n}}) = \vert \vert x^*_{\mathrm{b, n}} - x^r\vert \vert,
\end{equation}
where $x^r$ is the $x$ location of the true robust minimum. The first metric
is similar to the concept of regret from decision theory \citep{Blum2007,
Kaelbling1996} at the suggested $x^*_{\mathrm{b, n}}$. Regret measures how
much you lose out by running at $x^*_{\mathrm{b, n}}$ compared to $x^r$.
Regret is always nonnegative since by definition, $x^r = \argmin_{x
\in [0, 1]^d} g(x, \alpha)$. The second metric is the distance from 
$x^*_{\mathrm{b, n}}$ to $x^r$. For both, lower is better with 0 being
the floor if a method correctly identifies exactly $x^r$ as the 
BEAR or BOV.

\subsection{Empirical comparisons}
\label{sec:empcomp}

\subsubsection*{One-dimensional examples}

\begin{figure}[ht!]
	\centering
	\includegraphics[width=5.84cm,trim=5 20 10 50]{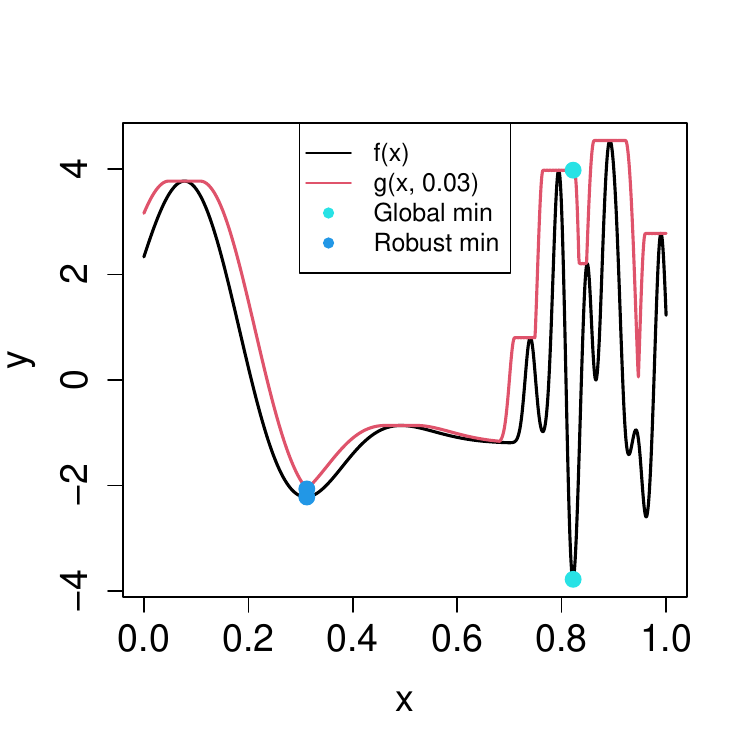}
	\includegraphics[width=5.84cm,trim=5 20 10 50]{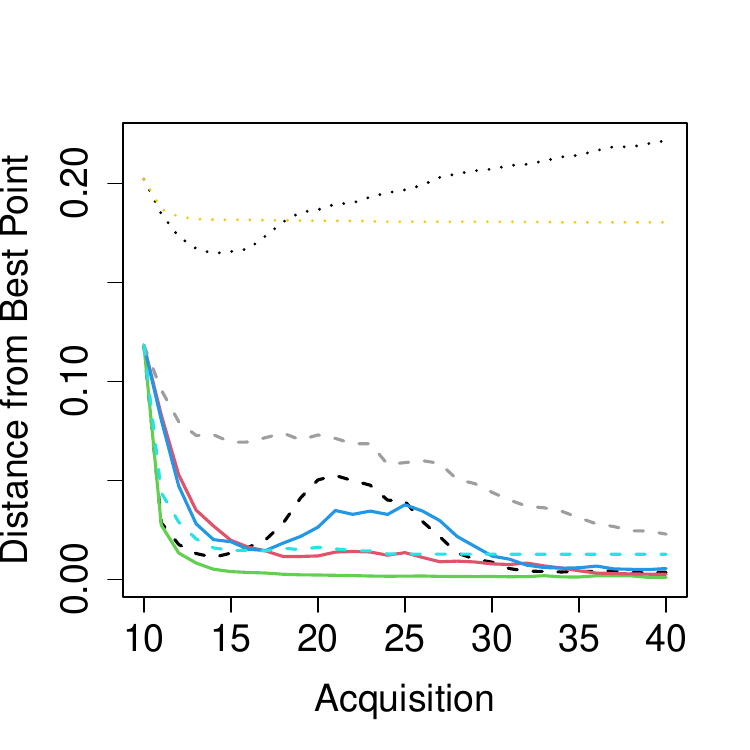}
	\includegraphics[width=5.84cm,trim=5 20 0 50]{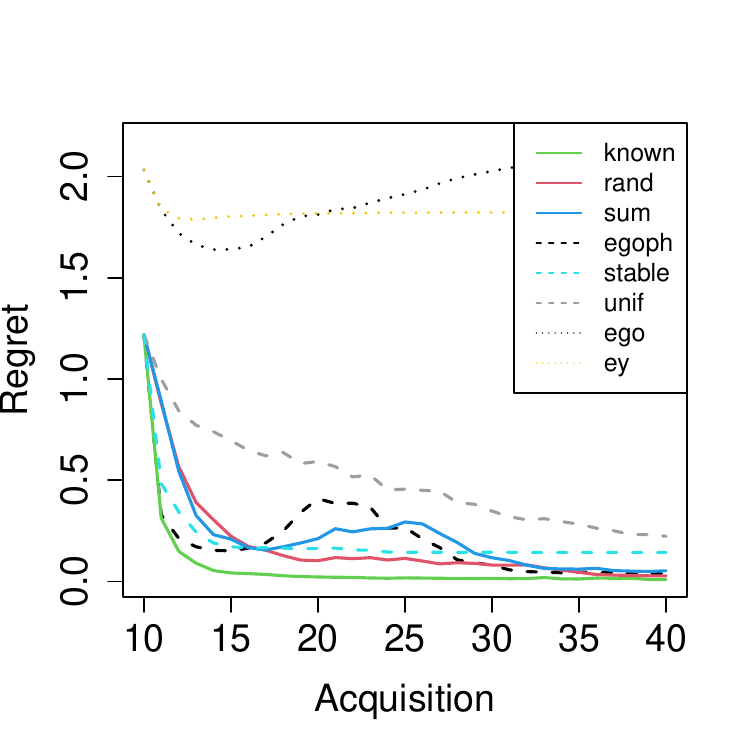}
	\caption{Left: The RKHS function and the robust surface with $\alpha = 
	0.03$. Middle: $d(x^*_{\mathrm{b, n}})$ after each acquisition. Right: 
	$r(x^*_{\mathrm{b, n}})$ after each acquisition.}
	\label{fig:rkhs}
\end{figure}

The left panel of Figure \ref{fig:rkhs} shows the 1d RKHS function used by
\cite{Assael2014}, and an adversary with $\alpha = 0.03$.  
Observe how $f$ has a smooth region for low $x$ values and a wiggly region for
high $x$. With $\alpha=0.03$, $x^r=0.312$ whereas $x^*=0.822$.  The adversary
is bumped up substantially nearby $x^*$ because the surface is so wiggly
there. Here we consider a final budget of $N=40$ with $\theta = 0.01$. Results
are provided in the middle and right panels of Figure \ref{fig:rkhs}. Observe
that REGO with known $\alpha$ performs best as it quickly gets close to $x^r$,
and more-or-less stays there. EGO with post hoc adversary does well at the
beginning, but after twenty acquisitions it explores around $x^*$ more,
retarding progress (explaining the ``bump'') toward $x^r$. ``Sum''-based REI
has a somewhat slighter bump.   Averaging over smaller $\alpha$-values favors
exploring the wiggly region. {\tt StableOPT} caps out at a worse solution than
any of the REI-based methods. Those based on BOV, i.e., \blu{conventional BO
methods} like ``ego'' and ``ey'', fare worst of all -- even worse than
``unif''.  \blu{We wish to emphasize that all other methods, i.e., besides
``ego'', ``ey'' and ``unif'', utilize adversarial reasoning.  All but {\tt
stableOPT} in that group deploy some number of elements comprising our novel
contribution, i.e., a post hoc adversary, or full REGO.}

\begin{figure}[ht!]
	\centering
	\includegraphics[width=5.84cm,trim=5 10 10 50]{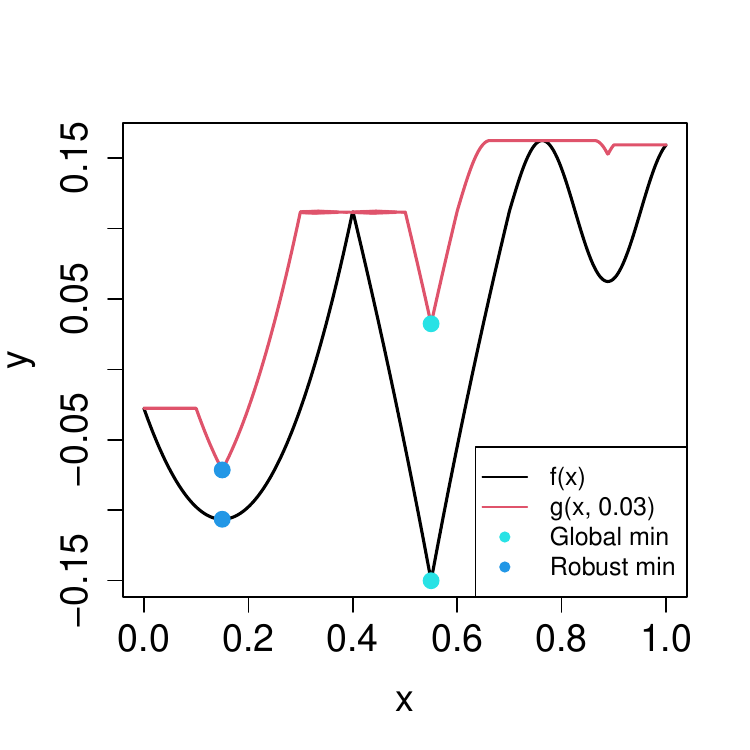}
	\includegraphics[width=5.84cm,trim=5 10 10 50]{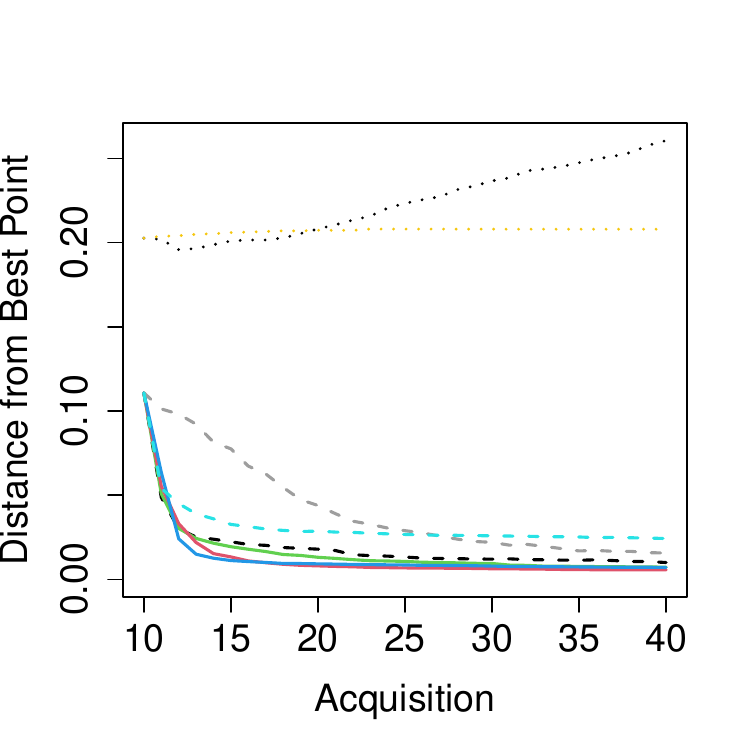}
	\includegraphics[width=5.84cm,trim=5 10 0 50]{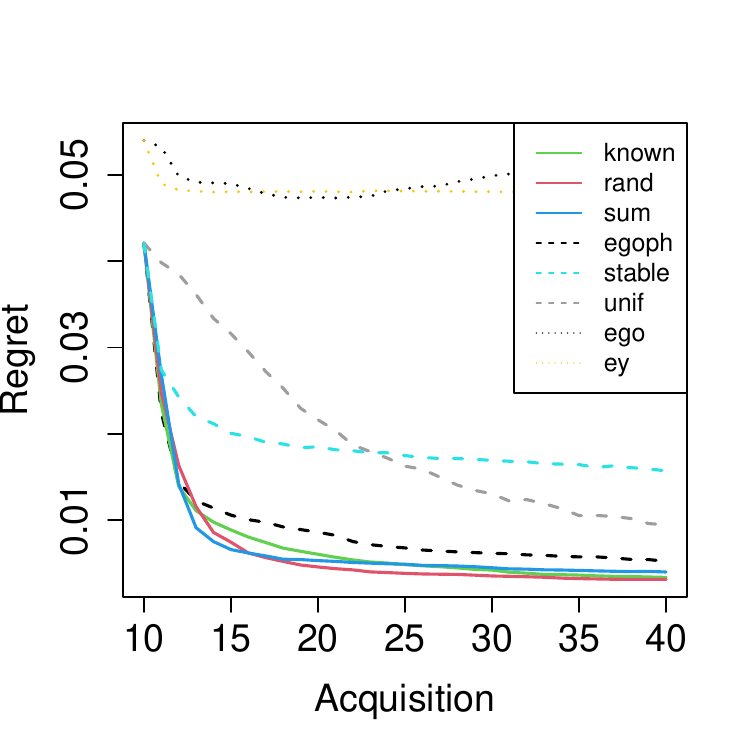}
	\caption{Left: The Figure \ref{fig:1D-Example} function and the robust 
		surface with $\alpha = 0.075$. Middle: $d(x^*_{\mathrm{b, n}})$ after
		each acquisition. Right: $r(x^*_{\mathrm{b, n}})$ after each
		acquisition.}
	\label{fig:ryan}
\end{figure}

Figure \ref{fig:ryan} shows a 1d test function of our own creation, first 
depicted in Figure \ref{fig:1D-Example}, defined as:
\begin{equation}
	\label{eq:ryan}
	f(x) =
		\begin{cases}
		3.5(x - 0.15)^2 + \log(1.3) & x < 0.4\\
		\log(1 + \vert 2(x - 0.55) \vert) & 0.4 \leq x < 0.7\\
		\frac{1}{20}\sin(25x - 17.5) + \log(1.3) & 0.7 \leq x.
		\end{cases}
\end{equation}
In this problem, $x^*=0.55$ and $x^r=0.15$ using $\alpha = 0.075$. All GP
surrogates used $\theta = 0.25$. We see a similar story here as for our first
test problem.  The only notable difference here is that EGO does not get
drawn into the peaky region, likely because it is less pronounced. EGO favors
sampling around $x^*$ initially, but eventually explores the rest of the input
space, including around $x^r$. Interestingly, REGO with known $\alpha$
performs a little worse than either of the unknown $\alpha$ methods.

\subsubsection*{Two-dimensional examples}

\begin{figure}[ht!]
	\includegraphics[width=6.6cm,trim=0 20 0 30]{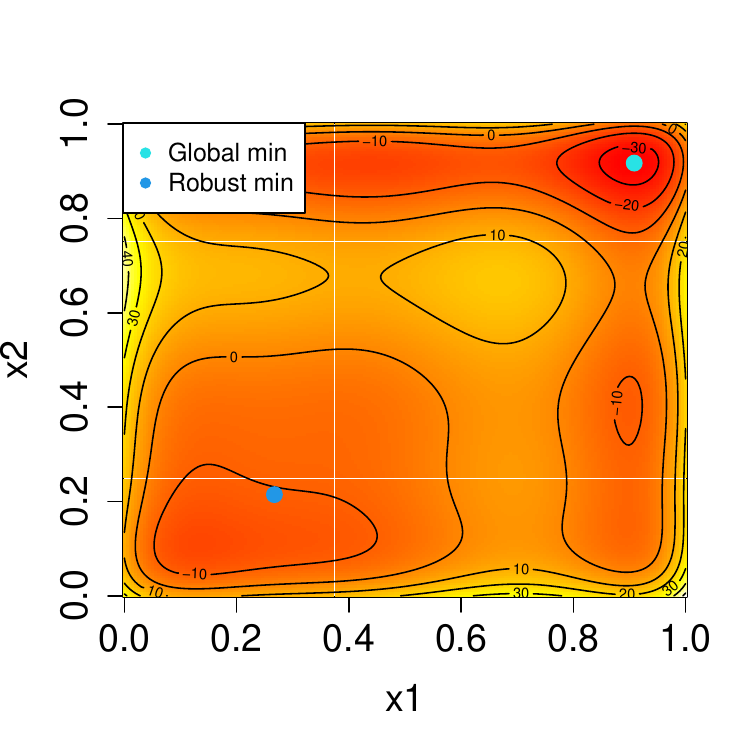}
	\includegraphics[width=6.6cm,trim=0 20 0 30]{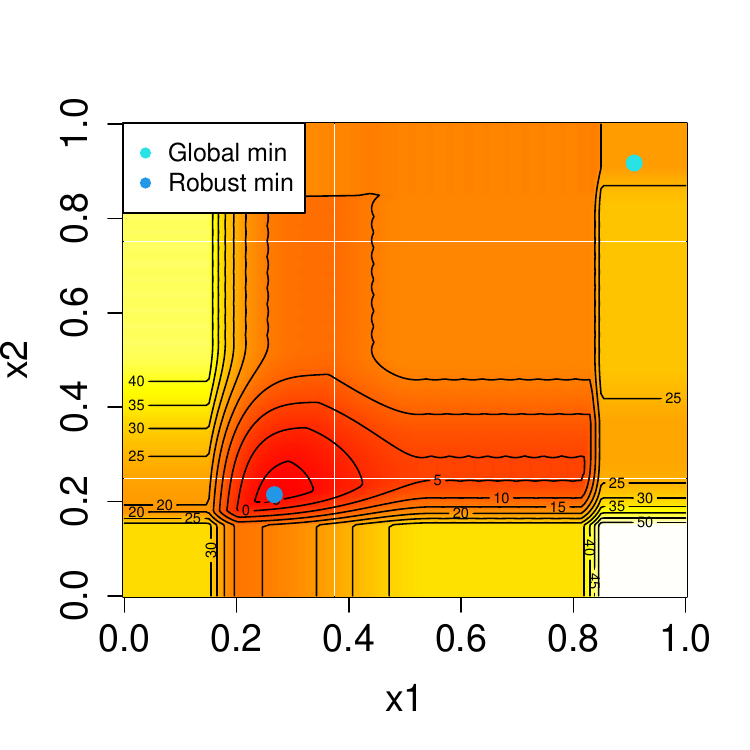}
	\includegraphics[width=6.6cm,trim=0 20 0 30]{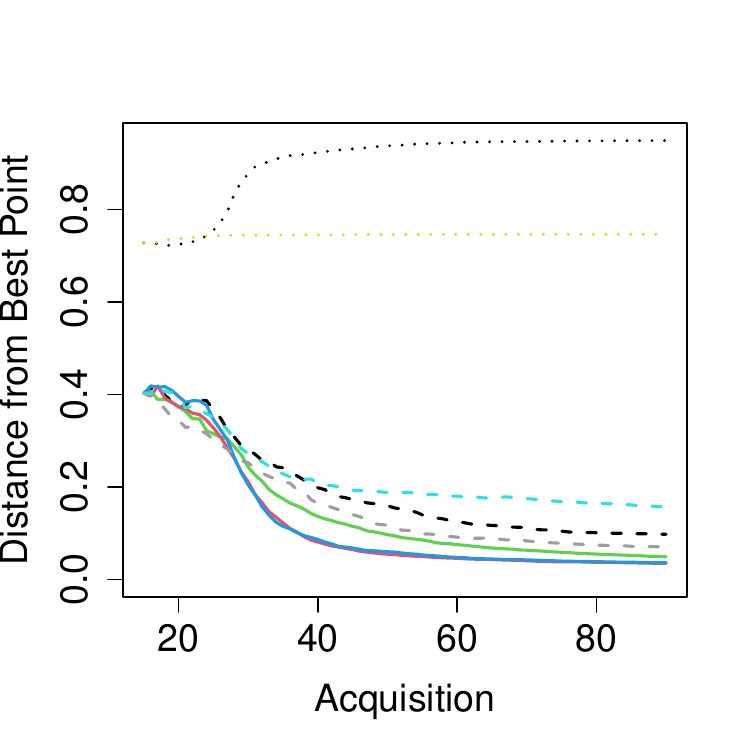}
	\includegraphics[width=6.6cm,trim=0 20 0 30]{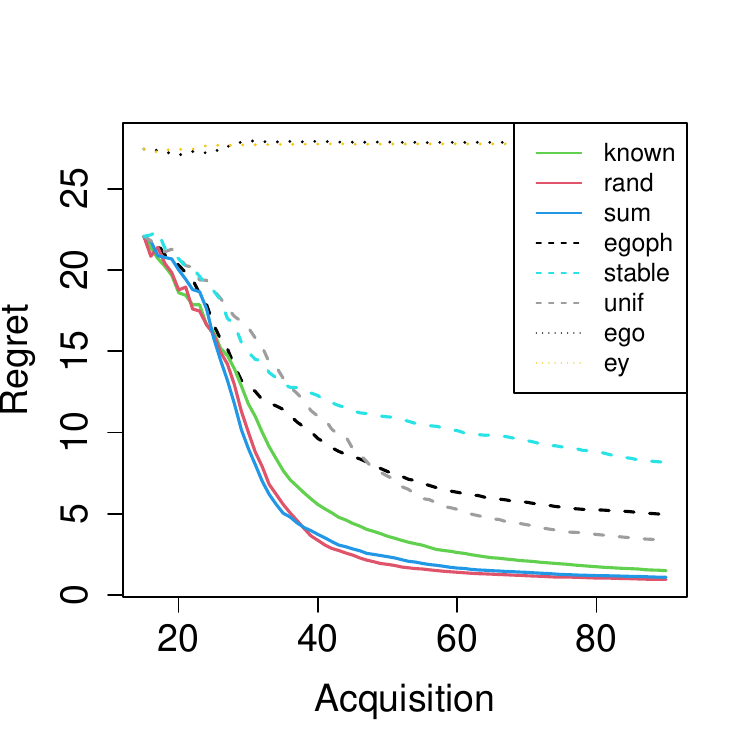}
	\caption{Top: The Bertsimas function on the left and the robust surface 
		with $\alpha = 0.15$ on the right. Bottom: $d(x^*_{\mathrm{b, n}})$ 
		(left) and $r(x^*_{\mathrm{b, n}})$ (right) after 
		each acquisition.}
	\label{fig:bertsima2d}
\end{figure}

The top-left panel of Figure \ref{fig:bertsima2d} shows a test problem
from \cite{Bertsimas2010}, \blu{a common test case in robust 
optimization, which is} defined as:
\begin{align}
	\label{eq:bert}
	f(x_1, x_2) = -2x_1^6 &+ 12.2x_1^5 - 21.2x_1^4 + 6.4x_1^3 + 4.7x_1^2 - 
	6.2x_1 - x_2^6 + 11x_2^5 - 43.3x_2^4\\
	&+74.8x_2^3 - 56.9x_2^2 + 10x_2 + 4.1x_1x_2 + 0.1x_1^2x_2^2 - 0.4x_1x_2^2 - 
	0.4x_1^2x_2.\nonumber
\end{align}
We negated this function compared to \citeauthor{Bertsimas2010}, who were
interested in maximization. Originally, it was defined on $x_1 \in [-0.95,
3.2]$ and $x_2 \in [-0.45, 4.4]$ with $x^*=(2.8, 4.0)$.  We coded inputs to
$[0, 1]^2$ so that the true minimum is at $(0.918, 0.908)$. In the scaled
space, $x^r = (0.2673, 0.2146)$ using $\alpha = 0.15$. We used $N=90$ and
$\theta=1.1$. \blu{Appendix \ref{app:esttheta} considers a variation where
$\hat{\theta}$ is re-estimated after each acquisition.  The results are
very similar, but noisier.} The bottom row of panels in Figure 
\ref{fig:bertsima2d} show that non-fixed $\alpha$ for REGO can give superior 
performance in early acquisitions. {\tt StableOPT} performs worse with this 
problem because the objective surface near $x^r$ is relatively more peaked, say 
compared to the 1d RKHS example.
Regret is trending toward 0 when using BEAR for acquisitions, with REGO-based
methods leading the charge. EGO with post hoc adversary performs much worse
for this problem. EGO-based acquisitions heavily cluster near $x^*$
which thwarts consistent identification of $x^r$ even with a post hoc surrogate.

\begin{figure}[ht!]
	\centering
	\includegraphics[width=6.6cm,trim=0 10 0 50]{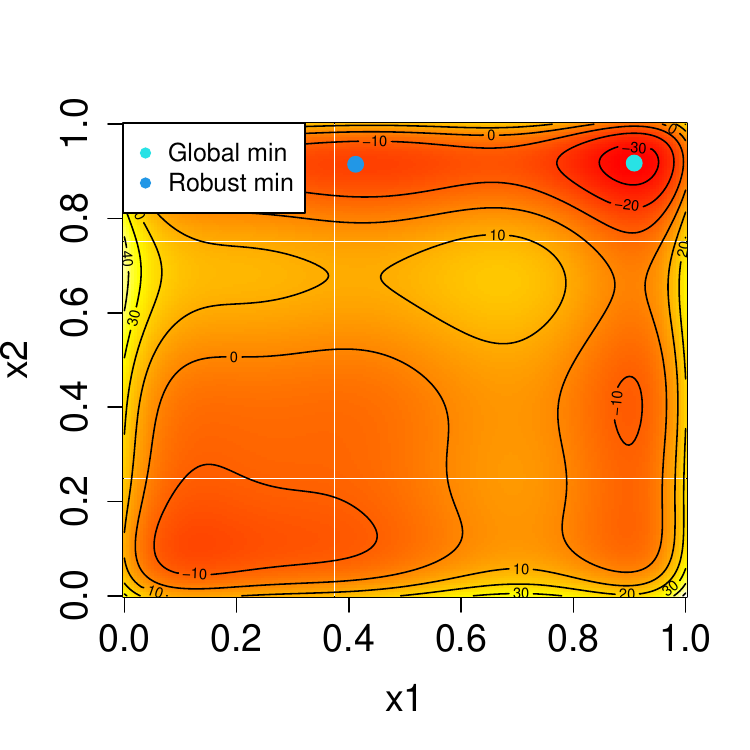}
	\includegraphics[width=6.6cm,trim=0 10 0 50]{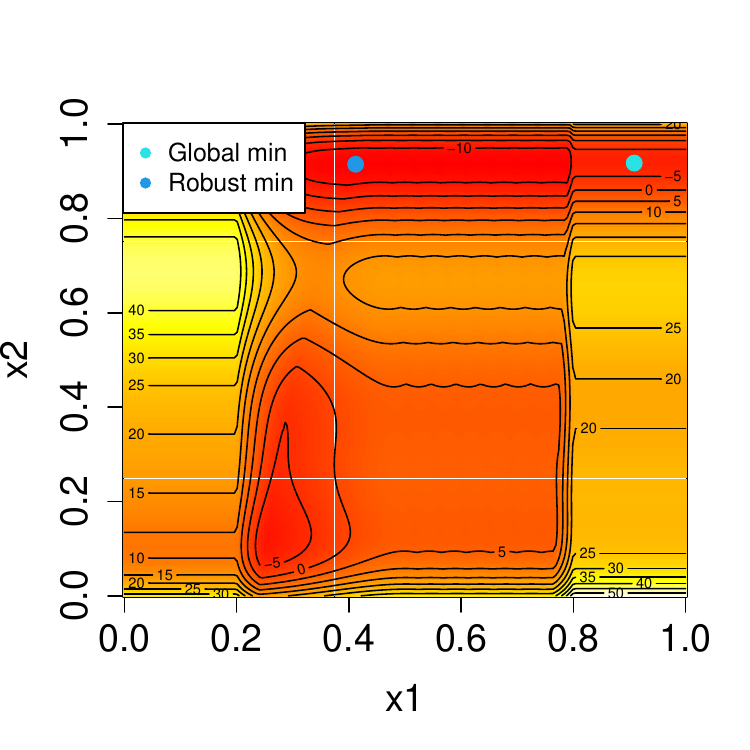}
	\includegraphics[width=6.6cm,trim=0 10 0 50]{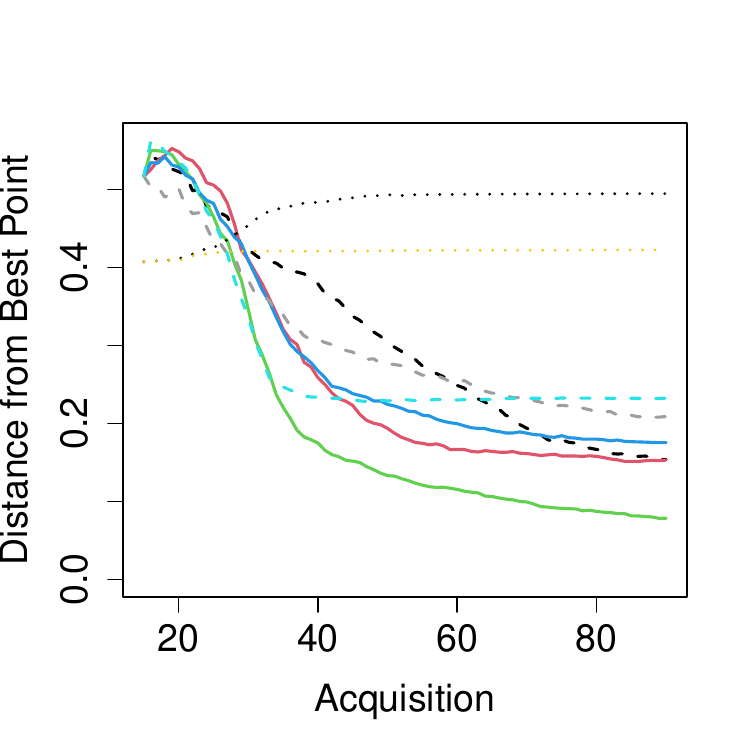}
	\includegraphics[width=6.6cm,trim=0 10 0 50]{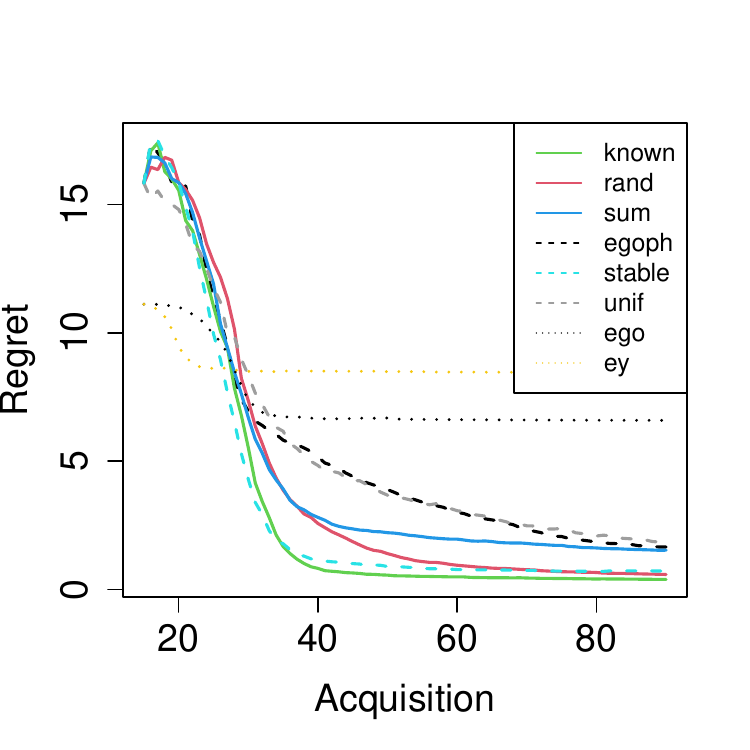}
	\caption{Top: The Bertsimas function on the left with the robust surface 
	with $\alpha = (0.2, 0)$ on the right. Bottom: $d(x^*_{\mathrm{b, n}})$ 
	(left) and $r(x^*_{\mathrm{b, n}})$ (right) 
	after each acquisition.}
\label{fig:bertsima1d}
\end{figure}

Figure \ref{fig:bertsima1d} considers the same test problem (\ref{eq:bert}),
except this time we use $\alpha = (0.2, 0)$, meaning no robustness required in
$x_2$. This moves $x^r$ to $(0.412, 0.915)$ in the scaled space. In this
problem, the robust surface is fairly flat meaning that when an algorithm
finds $x_2 = 0.915$, $x_1 \in [0.35, 0.75]$ gives a similar robust output. For
that reason, all of the methods perform worse when trying to pin down the
exact $x_1$ location, so by looking at metric $d(x^*_{\mathrm{b}})$ from
Eq.~(\ref{eq:MCmetrics}), all methods appear to be doing worse, \blu{whereas}
$r(x^*_{\mathrm{b}})$  goes to 0 relatively quickly. A shallower robust
minimum favors {\tt StableOPT}.  Since that comparator never actually
evaluates at $x^r$, \blu{here} it suffices to find $x_1 \in [0.35, 0.75]$,
which it does quite easily. Knowing true $\alpha$ for REI helps considerably.
This makes sense because omitting an entire dimension from robust
consideration is informative. Nevertheless, alternatives using random and
aggregate $\alpha$-values perform well.

\subsubsection*{Higher dimension}

Our final set of test functions comes from the Rosenbrock family 
\citep{Dixon1979},
\begin{equation}
	\label{eq:rosenbrock}
	\sum_{i = 1}^{d - 1} \left(100(x_{i +  1} - x_i)^2 + (x_i - 1)^2\right).
\end{equation}
defined in arbitrary dimension $d$. Originally in $[-2.48, 2.48]^d$,
we again scale to $[0, 1]^d$. Although our focus here will be on $d=4$,
visualization easier in 2d. \blu{Appendix \ref{app:highdim} provides results
for a 6d variation, which are quite similar.}
\begin{figure}[ht!]
	\centering
	\includegraphics[width=5.5cm,trim=0 63 0 
	50,clip=true]{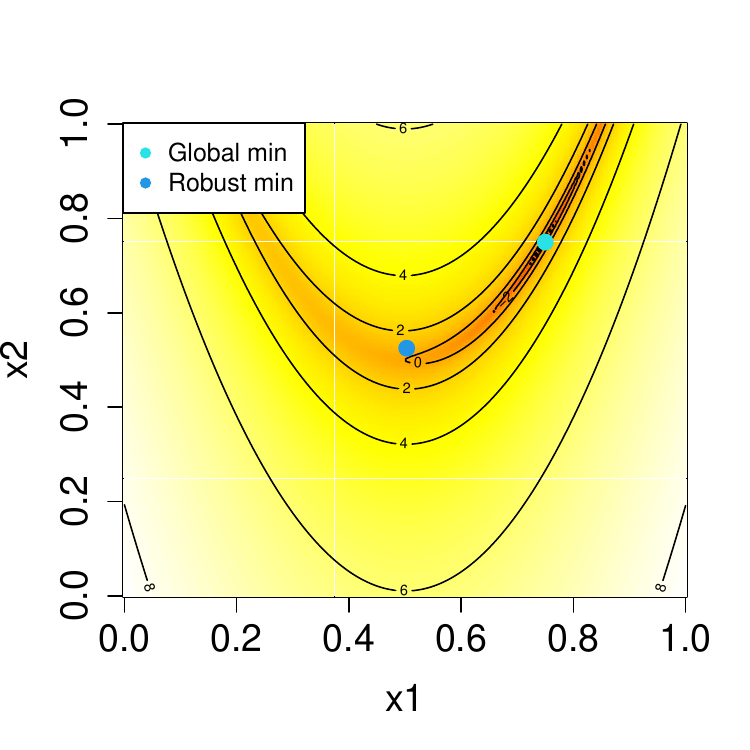}
	\includegraphics[width=5.5cm,trim=0 63 0 50,clip=true]{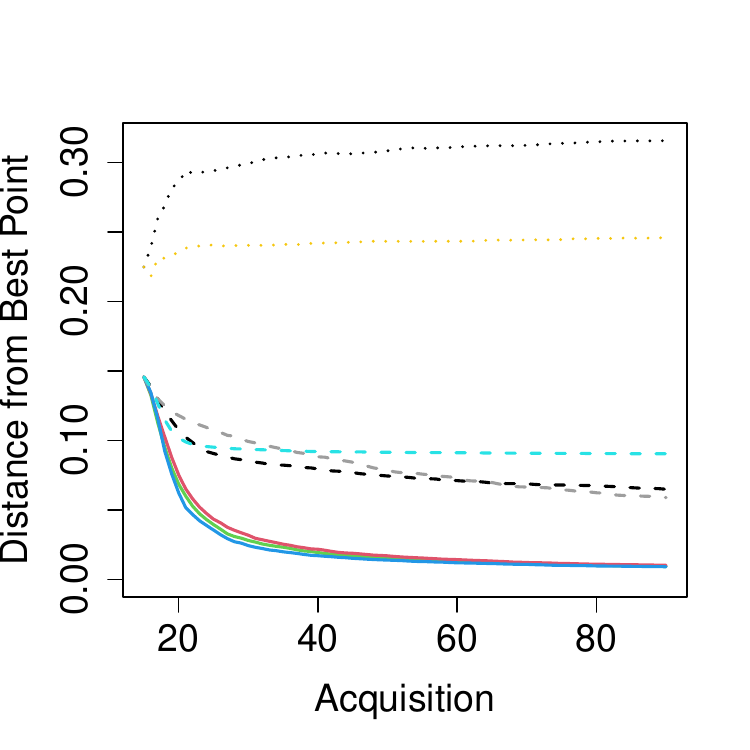}
	\includegraphics[width=5.5cm,trim=0 63 0 50,clip=true]{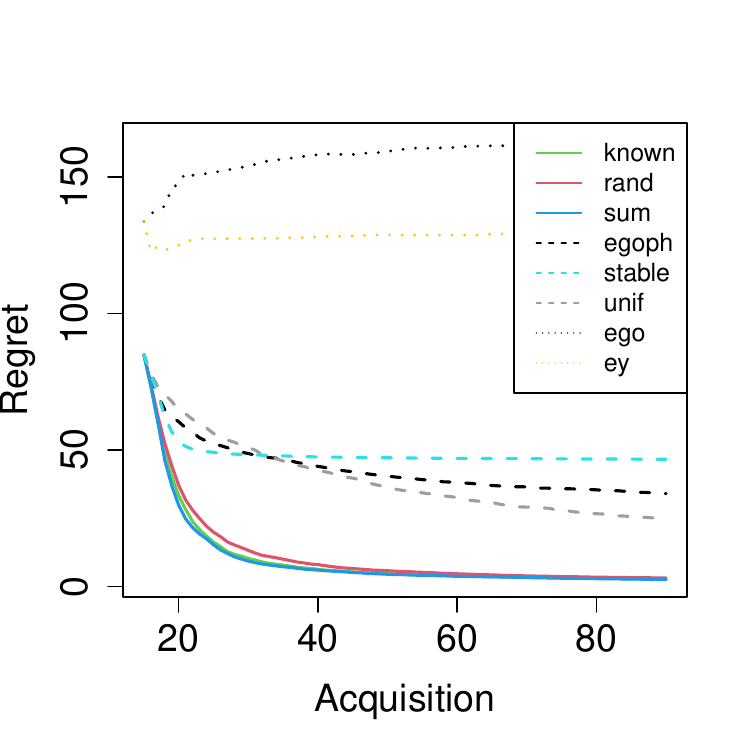}
	\includegraphics[width=5.5cm,trim=0 10 0 50]{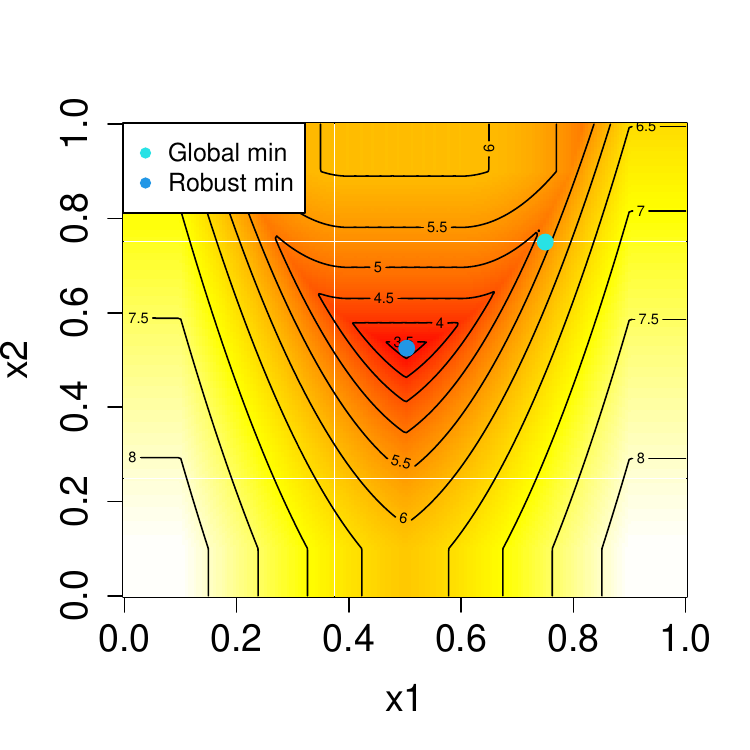}
	\includegraphics[width=5.5cm,trim=0 10 0 50]{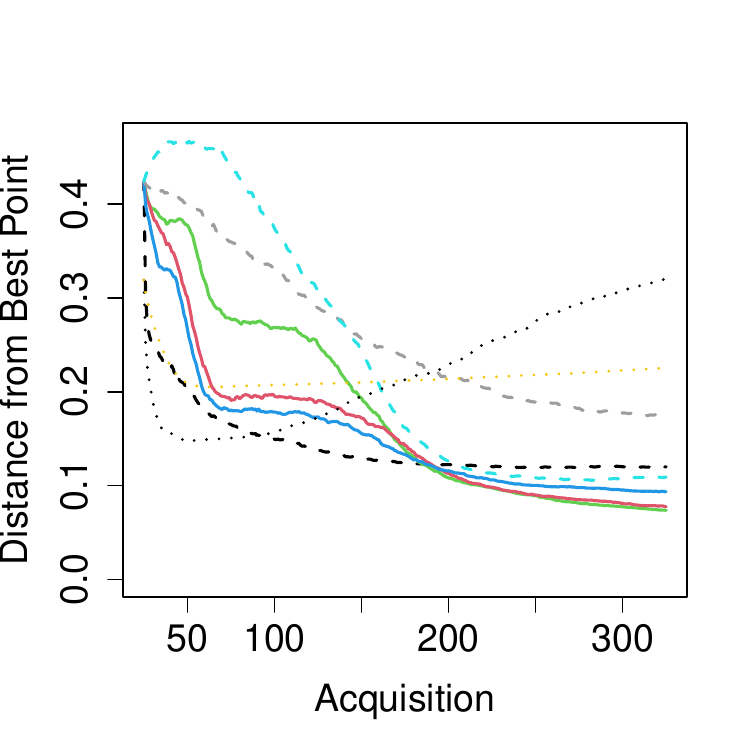}
	\includegraphics[width=5.5cm,trim=0 10 0 50]{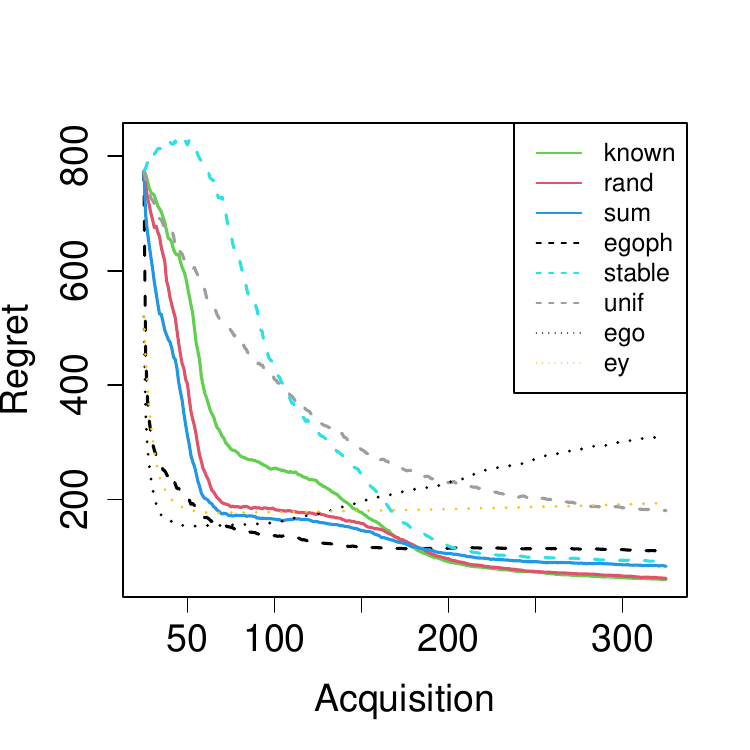}
	\caption{Left: 2d log Rosenbrock function on the top and the robust 
	surface with $\alpha = 0.1$ on the bottom. Top: 
	$d(x^*_{\mathrm{b, n}})$ in the middle and $r(x^*_{\mathrm{b, n}})$ 
	on the right for 2d. Bottom: similarly in 4d.}
	\label{fig:rosenbrock2d}
\end{figure}
The top-left panel of Figure \ref{fig:rosenbrock2d} shows 
 outputs on the log scale (2d). Here $x^*=(0.75, 0.75)$ and
 $x^r=(0.503, 0.525)$ when $\alpha = 0.1$.  A visual of this adversary is
 in the bottom-left panel. In 2d, we set $\theta=0.9$ and in 4d, we use
 $\theta = 0.05$. Results for 2d are in the top row (middle and right
 panels), and for 4d are in the bottom row, respectively. REGO shines in both
 cases because $x^*$ is in a peaked region and $x^r$ is in a shallow
 one. EGO with post hoc adversary does well early on, but stops improving
 much after about 150 acquisitions. {\tt StableOPT} has the same issues of
 sampling around $x^r$ that we have seen throughout -- never actually sampling
 it.

\subsection{Supplementary empirical analysis}
\label{sec:supp}

\begin{figure}[ht!]
	\includegraphics[width=5.2cm,trim=0 40 75 45,clip]{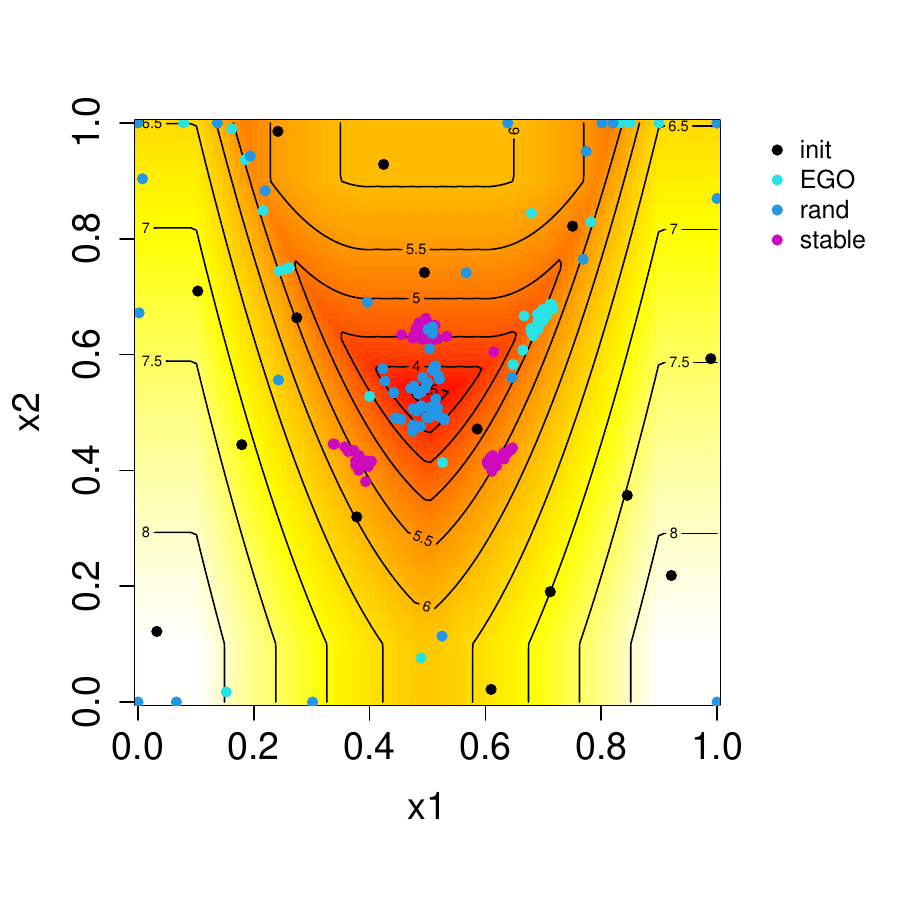}
	\includegraphics[width=5.2cm,trim=0 40 75 45,clip]{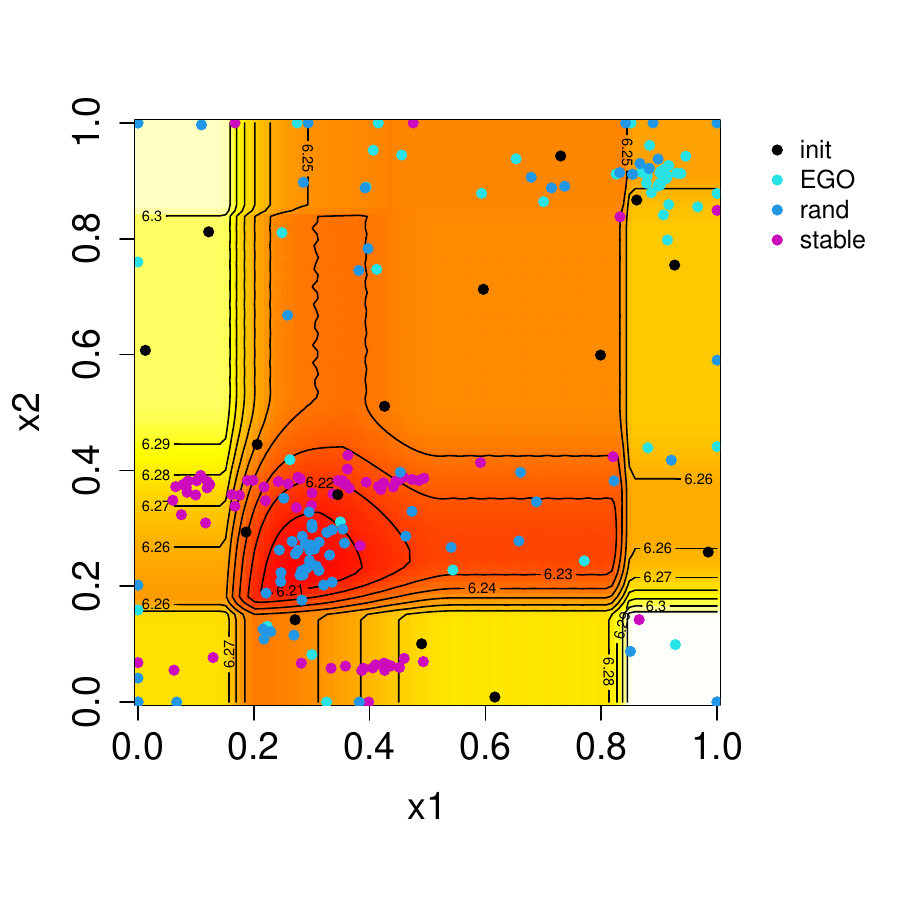}
	\includegraphics[width=6.3cm,trim=0 40 0 45,clip]{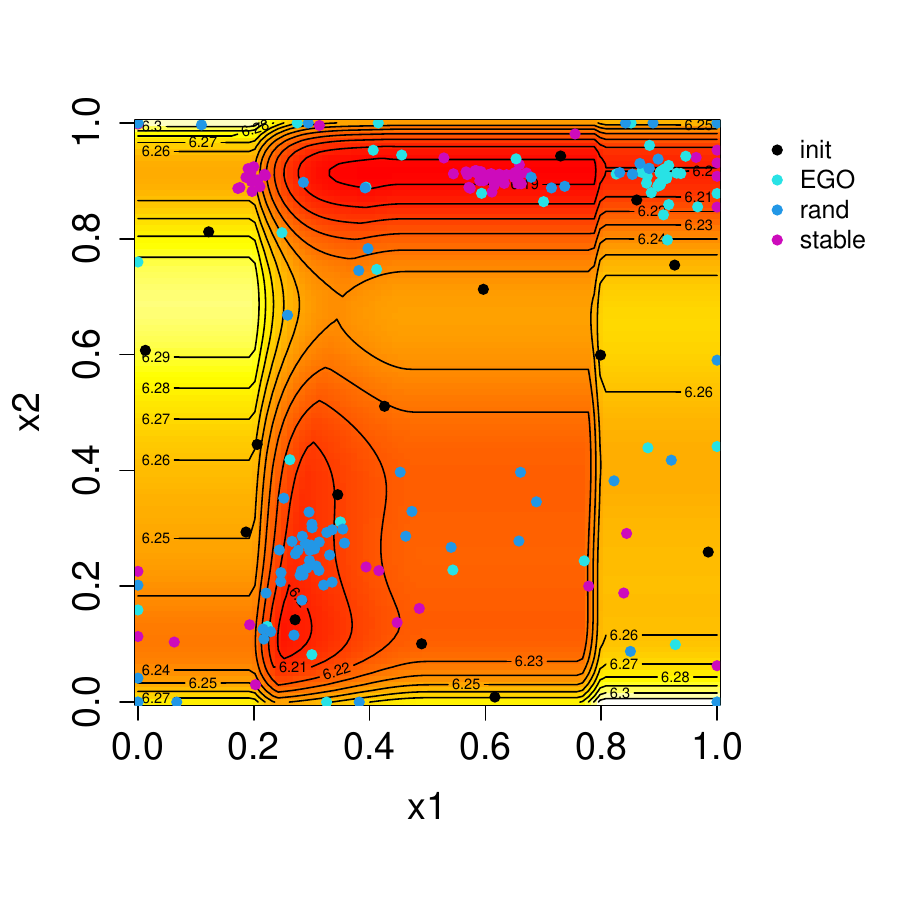}
	\caption{Sample acquisitions for EGO, REGO with random $\alpha$ and 
	{\tt StableOPT} for 2d Rosenbrock with $\alpha = 0.1$ (left), and Bertsimas
	functions with $\alpha = 0.15$ (middle), and $\alpha = (0.2, 0)$ (right).}
	\label{fig:samps}
\end{figure}

An instructive, qualitative way to evaluate each acquisition algorithm is to
inspect the final collection of samples (at $N$), to see visually if things
look better for robust variations. Figure \ref{fig:samps} shows the
final samples of one representative MC iteration for EGO, REGO with random
$\alpha$ and {\tt StableOPT} for 2d Rosenbrock (\ref{eq:rosenbrock})
(left panel) and both Bertsimas (\ref{eq:bert}) variations (middle
and right panels).

Consider Rosenbrock first.  Here, EGO has most of its acquisitions in a mass
around $x^*$ with \blu{a few} dispersed throughout the rest of the
space. This is exactly what EGO is designed to do: target the global minimum,
but still explore other areas. REGO has a similar amount of space-fillingness,
but the target cluster is focused on $x^r$ rather than $x^*$. On the other
hand, {\tt StableOPT} has almost no exploration points. Nearly all of its
acquisitions are on the perimeter of a bounding box around $x^r$. While {\tt
StableOPT} does a great job of picking out where $x^r$ is, intuitively we
do not need 70+ acquisitions all right next to each other. Some of those
acquisitions could better facilitate learning of the surface by exploring
elsewhere.

Moving to the Bertsimas panels of the figure, similar behavior may be
observed. REGO and EGO have some space-filling points, but mostly target $x^r$
for REGO and $x^*$ for EGO. {\tt StableOPT} again puts almost all of its
acquisitions near $x^r$ with relatively little exploration. But the main
takeaway from the Bertsimas plots is that, since REGO does not require setting
$\alpha$ beforehand, it gives sensible designs for multiple $\alpha$ values
(the blue points are the exactly the same in both panels). 
Looking more closely at the REGO design, observe
that all three minima (global and robust with $\alpha = 0.15$ and $\alpha =
(0.2, 0)$) have many acquisitions around them. This shows the power of REGO,
capturing all levels of $\alpha$ and allowing the user to delay specifying
$\alpha$ until after experimental design.
\begin{figure}[ht!]
	\centering
	\vbox{\vspace{0cm}
	\includegraphics[width=5.5cm,trim=0 65 0 30,clip=true]{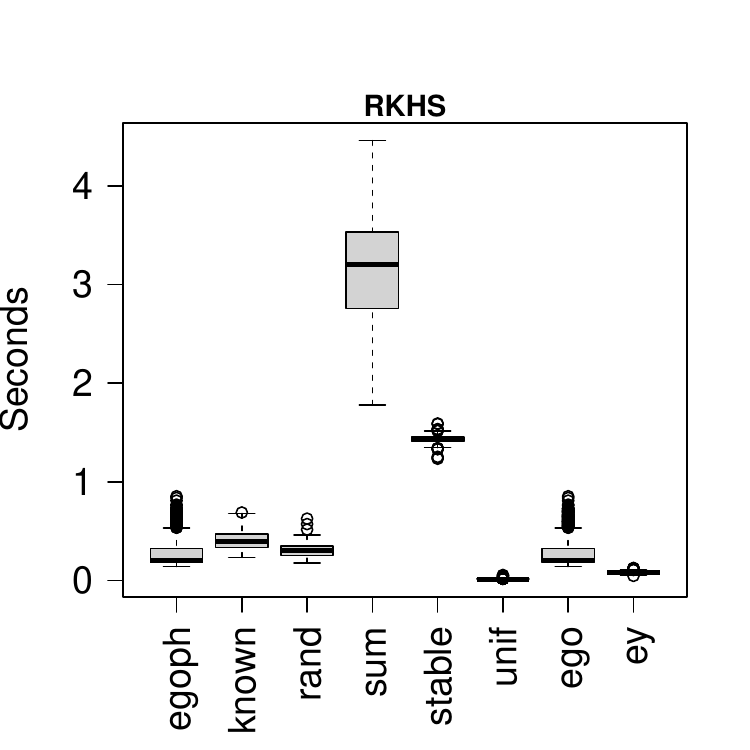}
	\includegraphics[width=5.5cm,trim=0 65 0 30,clip=true]{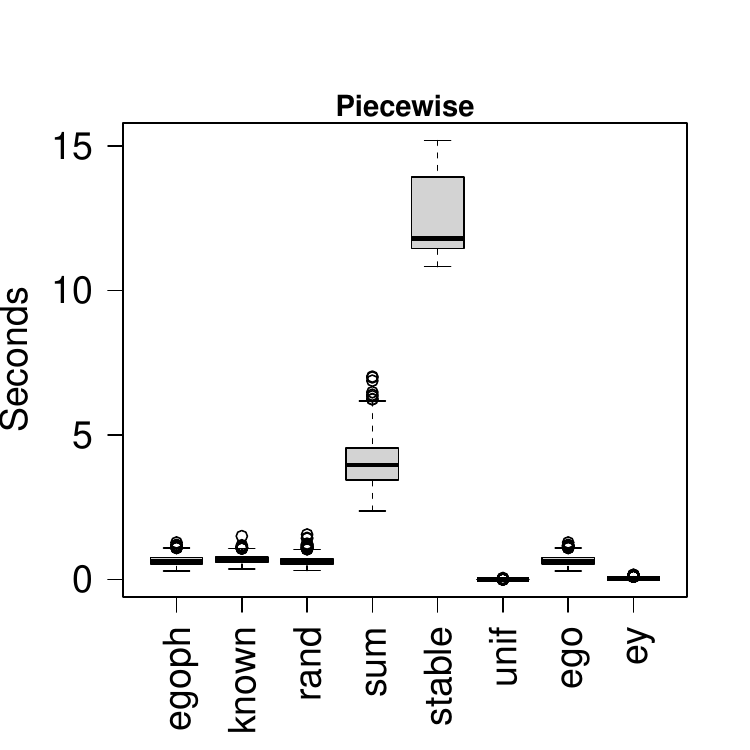}
	\includegraphics[width=5.5cm,trim=0 65 0 
	30,clip=true]{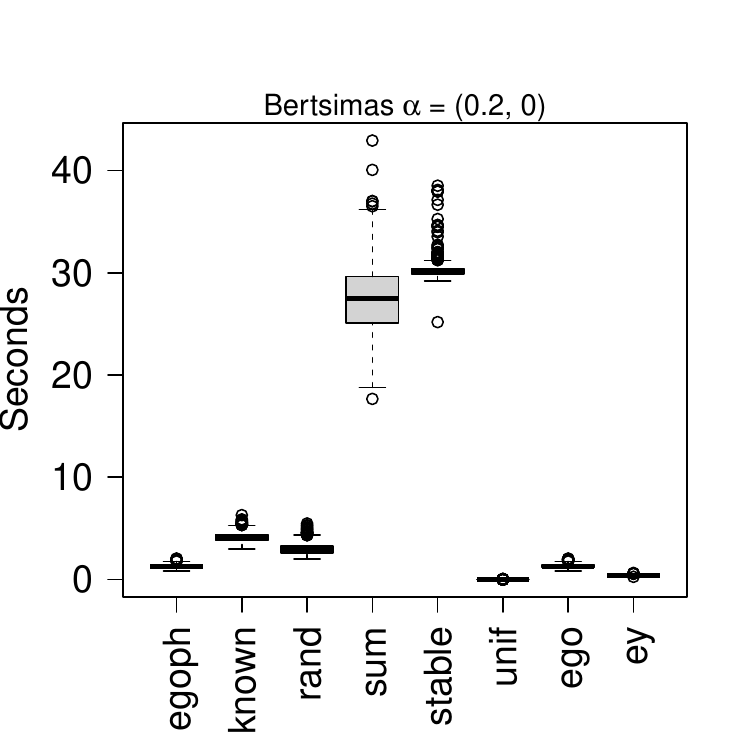}}
	\includegraphics[width=5.5cm,trim=0 0 0 30]{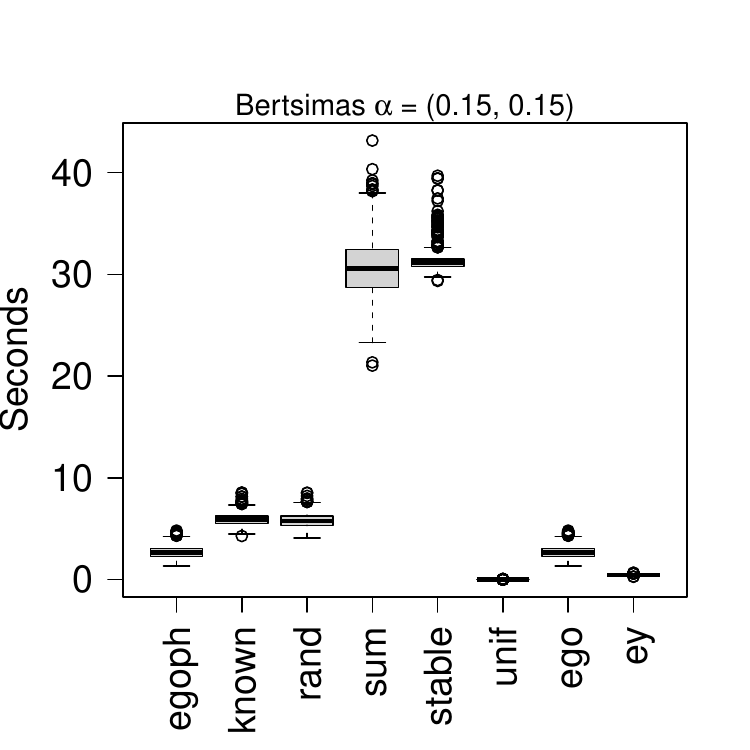}
	\includegraphics[width=5.5cm,trim=0 0 0 30]{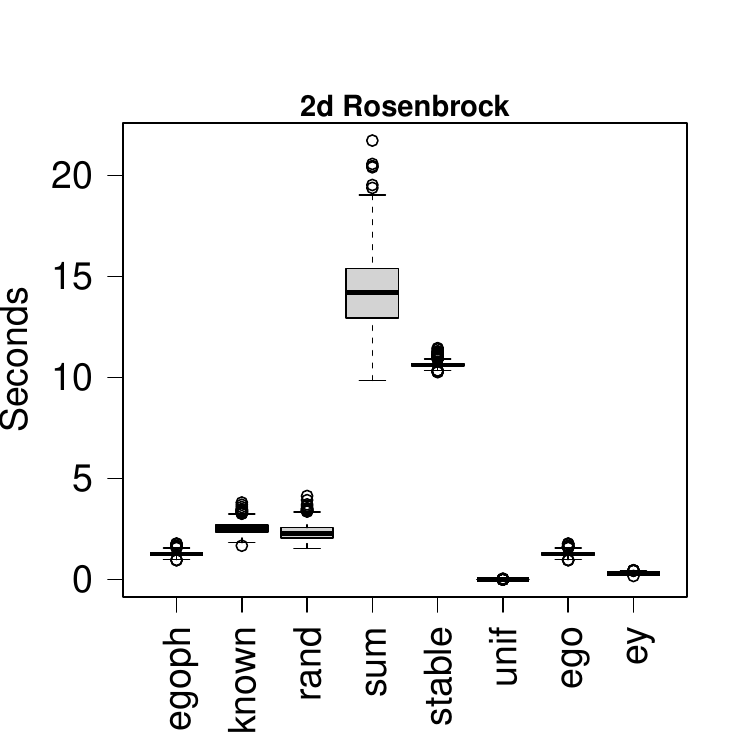}
	\includegraphics[width=5.5cm,trim=0 0 0 30]{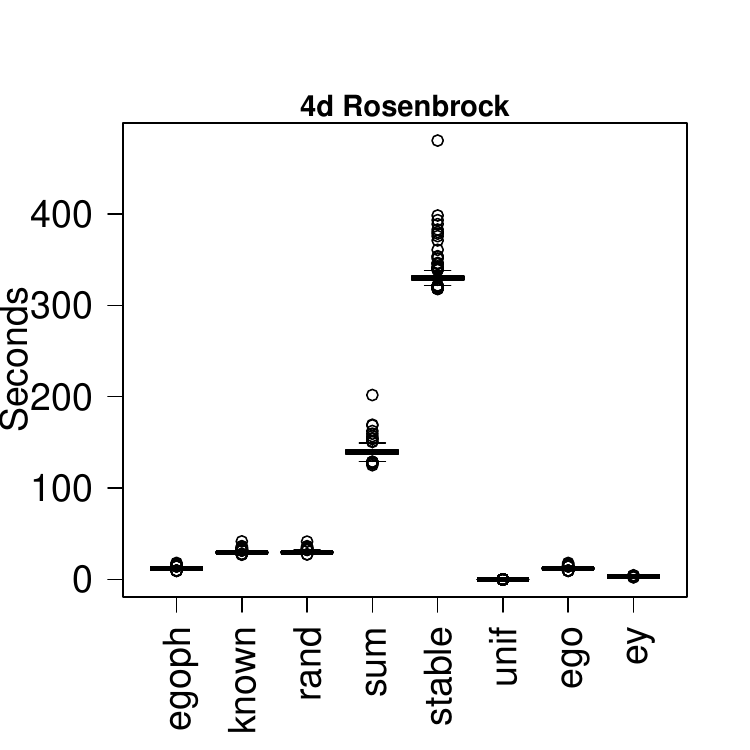}
	\caption{\blu{Cumulative timings for each method/test function.}}
	\label{fig:times}
\end{figure}

\blu{Timings for each method/problem are in Figure \ref{fig:times}.}
Comparators ``egoph'' and ``ego'' report identical timings because they
involve the same EI acquisition function. As you might expect, ``unif'' and EY
have the lowest times because they do less work. For the more competitive
methods, \blu{REGO is generally a little slower than EGO, but often faster
than {\tt StableOPT}}. 
\blu{The main bottleneck in BO is the $\mathcal{O}(n^3)$ matrix
decomposition(s) required for GP likelihood evaluation and prediction.  This $n$
is changing throughout acquisitions, and there is a different final $N$ in
each expariment.  Consequently we have provided cumulative timings in Figure
\ref{fig:times}.  Since the same underlying GP implementation is shared among
all of the competitors in our study, the timings are similar (within the same
test problem), with exceptions being  ``sum'' (aggregating many GP fits),
``unif'' (not requiring a GP), and ``stable'' (a totally different approach).
When amortizing over all $N$ acquisitions, it is clear that each takes just
mere seconds even in the worst of cases.  We think this is fast enough for
most real-world black-box evaluations typically involved in BO enterprises.}

\section{Robot pushing}
\label{sec:rob}

Here we consider a real world example to measure the effectiveness of REI.
The robot pushing simulator 
\citep[][https://github.com/zi-w/Max-value-Entropy-Search]{Kaelbling2017} has 
been used previously to test ordinary BO \citep{Wang2017a, Wang2017b} and robust 
\citep{Bogunovic2018} BO methodology. The simulator models a robot hand, with up 
to 14 tunable parameters, pushing a box with the goal of minimizing 
distance to a target location. 
Following \cite{Wang2017b}, we consider varying four of the tunable parameters, 
as detailed in Table \ref{tab:push_vars}.
\begin{table}[ht!]
	\centering
	\begin{tabular}{|c | c | c|} 
		\hline
		Parameter & Role & Range \\
		\hline\hline
		$r_x$ & initial x-location & $[-5, 5]$\\ 
		\hline
		$r_y$ & initial y-location & $[-5, 5]$\\
		\hline
		$r_\theta$ & initial angle & $[-\frac{\pi}{2}, 		
		\frac{\pi}{2}]$\\
		\hline
		$t_r$ & pushing strength & $[1, 30]$\\
		\hline
	\end{tabular}
	\caption{Parameters for the robot pushing simulator.}
	\label{tab:push_vars}
\end{table}
\noindent This simulator is coded in \textsf{Python} and uses an engine 
called \texttt{Box2D} \citep{Catto2011} to simulate the physics of pushing. 
Also following \cite{Wang2017b}, we consider two cases: one with a 
fixed hand angle, always facing the box, determined to be $r_\theta = 
\arctan(\frac{r_y}{r_x})$; and the other allowing for all 4 parameters to 
vary.  These create 3d and 4d problems that we call ``push3'' and ``push4'', 
respectively

We consider two further adaptations. First, we de-noised the simulator, so
that it is deterministic, which is more in line with our previous examples.
Second, rather than look at a single target location, we take the minimum
distance to two, geographically distinct target locations under squared and
un-squared distances respectively. We do this in order to manifest a version
of the problem that would require robust analysis. Having the box pushed the
full distance toward either target, minimizing the objective, yields an
output of 0 since $0^2 = 0$. However, the minimum around the unsquared target
will be shallower because the unsquared surface increases slower when the 
distance to the target is greater than 1. Robust BO prefers exploring the 
unsquared minimum while an ordinary, non-robust method would show no 
preference. Similarly, a BOV performance metric is indifferent to the target 
locations, while BEAR would favor the unsquared target.

For both ``push3'' and ``push4'' variations, we fixed the target locations at
$(-3, 3)$ and $(3, -3)$ with the latter being \blu{the squared target}. Since
the unsquared location is in the top-left quadrant, the optimal robust
location involves starting the robot hand in the bottom-right so that it
pushes the box up and left. Furthermore, because the robot cannot perfectly
control the initial hand location, if it is close to the origin, a minor
change in $r_x$ or $r_y$ leads to the hand pushing away from the target
location. For ``push3'' we use $\alpha = 0.1$, $x^r = (5, -5, 25.4)$, meaning
the robot hand starts as far in the bottom-right as possible and pushes quite
hard. The true setting $g(x^r, 0.1) = 1.37$ is lower than analogously pushing
toward the squared target location, $g((-5, 5, 25.4), 0.1) = 1.88$, solely due
to squaring as $1.37^2 \approx 1.88$.

\begin{figure}[ht!]
	\centering
	\includegraphics[scale=0.47,trim=0 10 0 60]{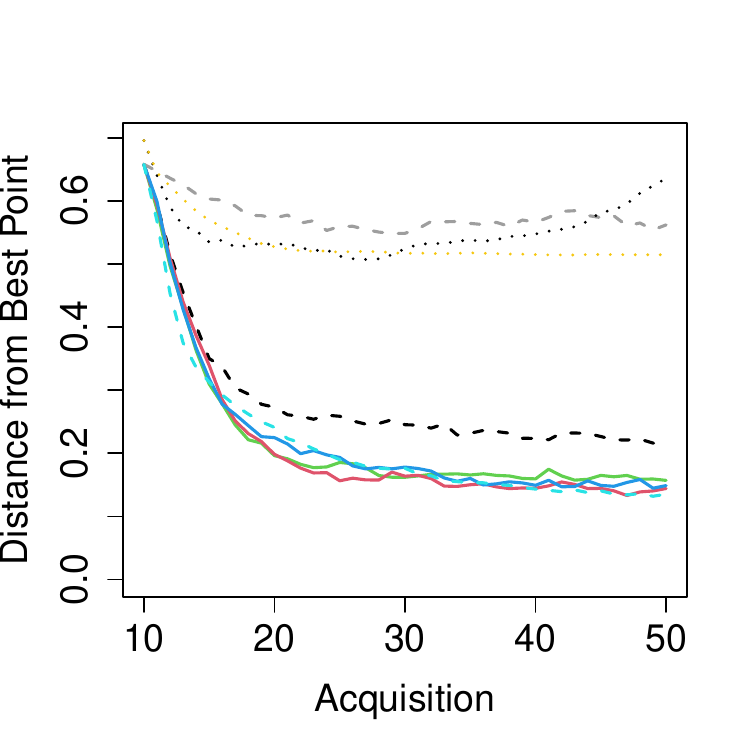}
	\includegraphics[scale=0.47,trim=25 10 0 60]{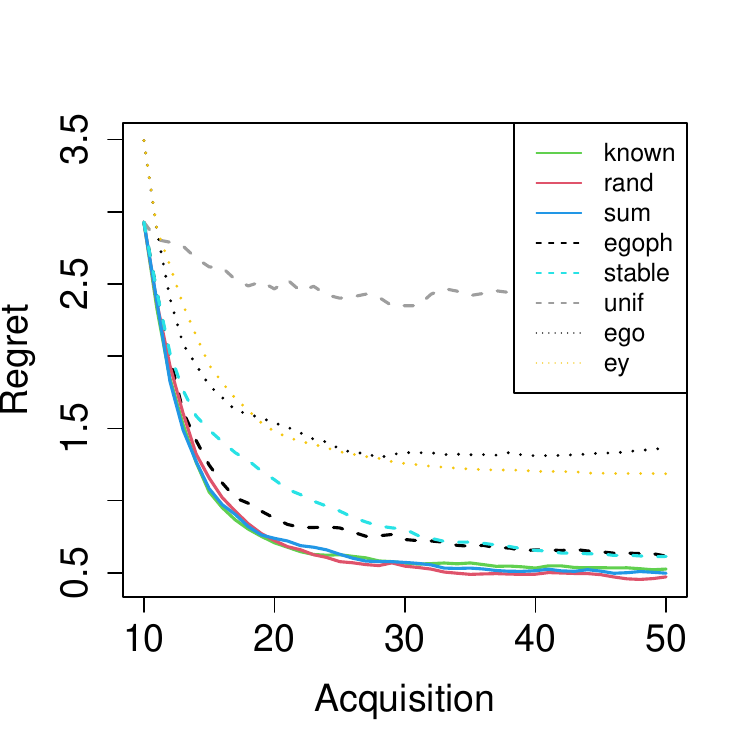}
	\includegraphics[scale=0.47,trim=25 10 0 60]{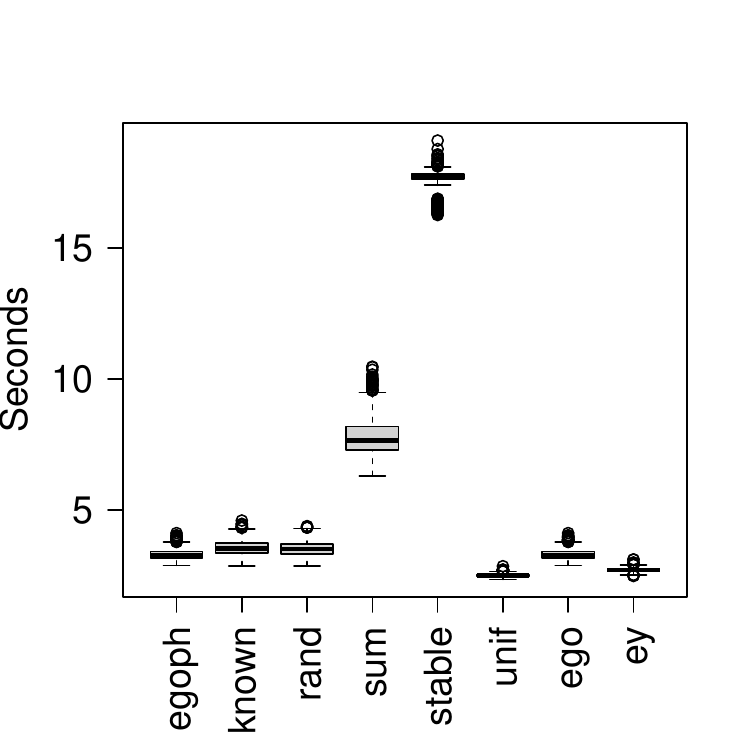}
	\caption{Results for ``push3'' $\alpha = 0.1$:
	$d(x^*_{\mathrm{b, n}})$  (left), $r(x^*_{\mathrm{b, 
	n}})$ (middle) and cumulative time (right).}
	\label{fig:push3}
\end{figure}

We compare each of the methods from Section \ref{sec:empcomp} in a similar
fashion by initializing with an LHS of size $n_0 = 10$ and acquiring forty
more points ($N=50$), repeating for 1,000 MC samples. Figure \ref{fig:push3}
summarizes our results. Observe that all of REI variations outperform the
others by minimizing regret faster and converging at a lower value. EI with
post hoc adversary and {\tt StableOPT} perform fairly well but suffer from
drawbacks similar to those described in Section \ref{sec:empcomp}. They cannot 
accommodate squared and unsquared target locations differently.

\begin{figure}[ht!]
	\centering
	\includegraphics[width=5.8cm,trim=25 10 0 0]{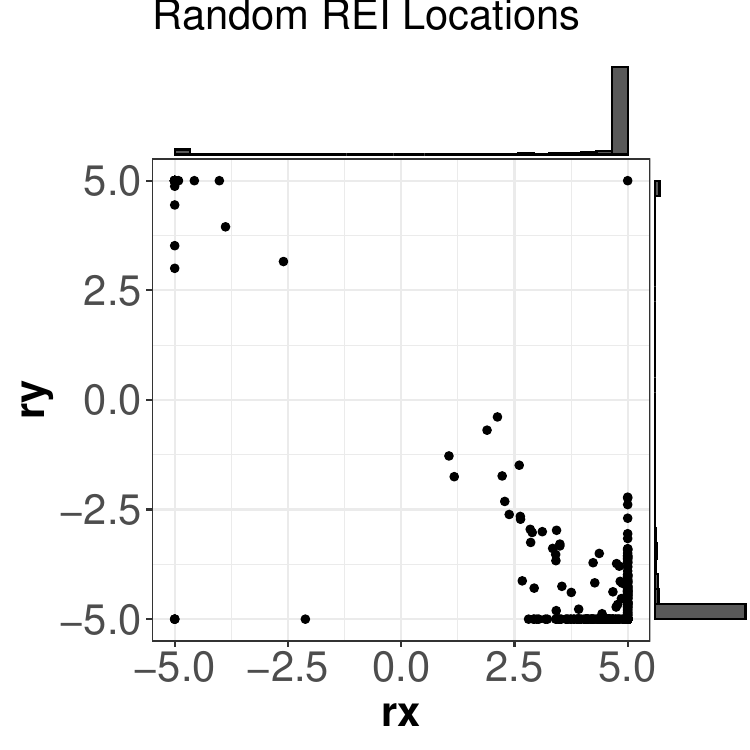}
	\includegraphics[width=5.8cm,trim=25 10 0 0]{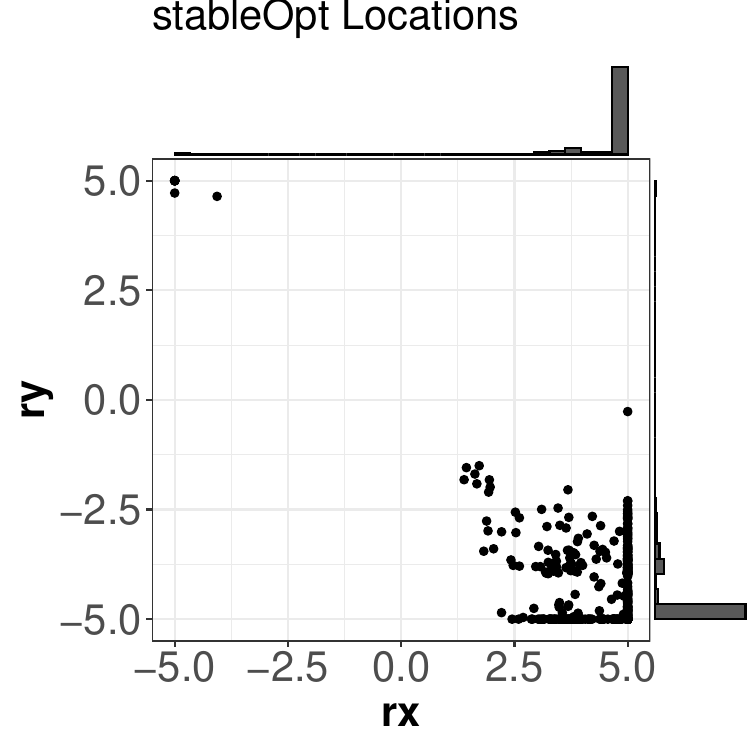}
	\includegraphics[width=5.8cm,trim=25 10 0 0]{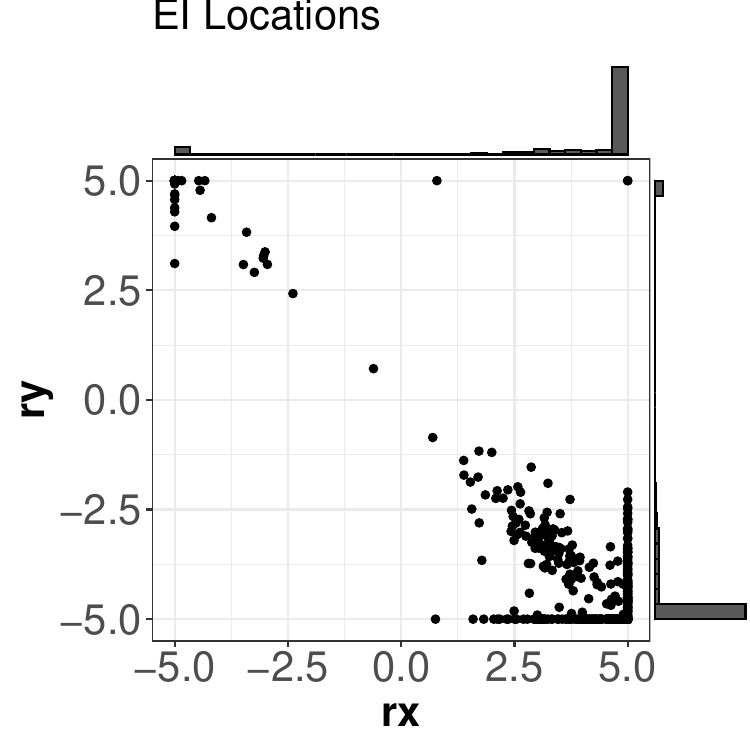}
	\caption{Distribution of $r_x$ and $r_y$ from $x^*_{\mathrm{bear}}$ for the 
	push3 problem after the final acquisition for ``rand'' REI using $\alpha = 
	0.1$ (left), {\tt StableOPT} (middle), and EI (right).}
	\label{fig:push3_hists}
\end{figure}

It is worth noting that, while REI performs well, the left panel of Figure
\ref{fig:push3} suggests that none of the methods are getting very close to
always finding $x^r$ when $d(x^*_{\mathrm{b, n}}) = 0$. (Distances are not
converging to zero.) To dig a little deeper, Figure \ref{fig:push3_hists}
explores $r_x$ and $r_y$ from $x^*_{\mathrm{bear}}$ for ``rand'' REI, {\tt
StableOPT} and EI with post hoc adversary. Notice that it is rare for any of
these comparators to push toward the ``wrong'' target location, i.e., finding
$(-5, 5)$ rather than $(5, -5)$ which would result in large $d(x^*_{\mathrm{b,
n}})$ and low $r(x^*_{\mathrm{b, n}})$. However, EI is attracted to the peaked
minimum more often than either of the other methods. REI has this problem
slightly more often than {\tt StableOPT}. However, {\tt StableOPT} and EI do
recommend $x^*_{\mathrm{bear}}$ in the bottom-right, but not all the way to
$(5, -5)$.  This happens more often than with REI. Only occassionally does REI
miss entirely, leading to large $d(x^*_{\mathrm{b, n}})$. On the other hand,
{\tt StableOPT} identifies the correct area slightly more often, but struggles
to pinpoint $x^r$. We conclude that both REI and {\tt StableOPT} perform well
-- much better than ordinary EI -- and any differences are largely a matter of
taste or tailoring to specific use cases, modulo computational considerations
(right panel).

\begin{figure}[ht!]
	\centering
	\includegraphics[scale=0.47,trim=0 10 0 50]{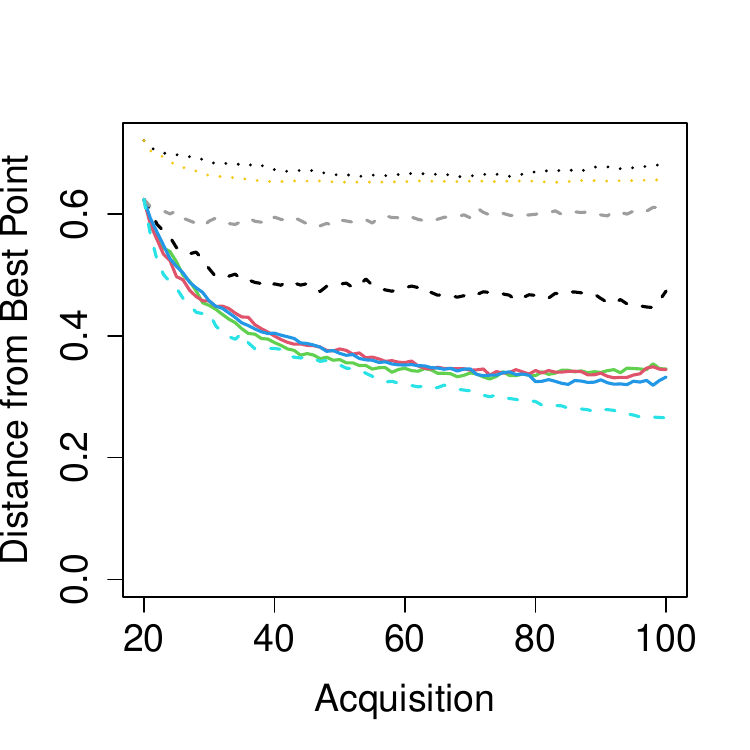}
	\includegraphics[scale=0.47,trim=25 10 0 50]{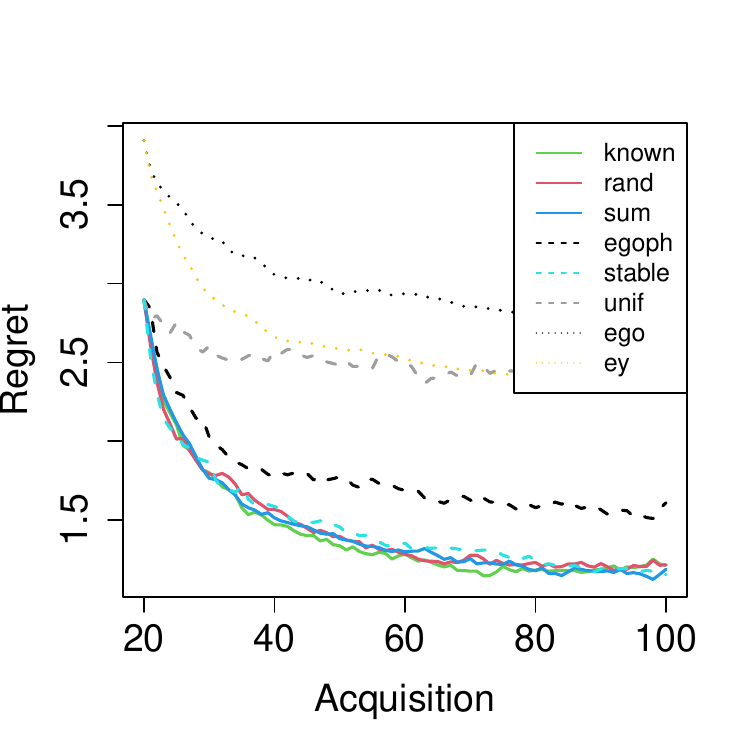}
	\includegraphics[scale=0.47,trim=25 10 0 50]{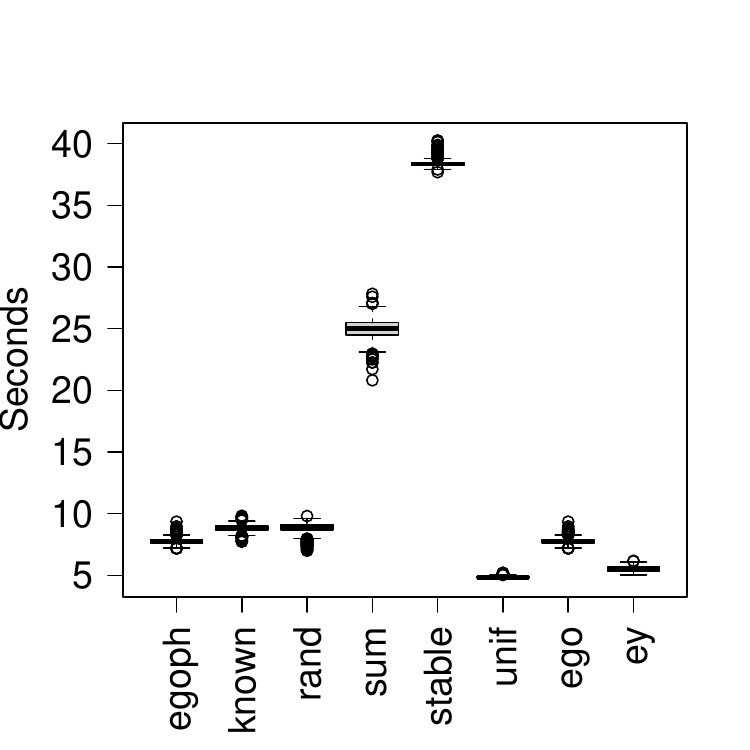}
	\caption{Results for ``push4'' $\alpha = 0.1$:
	$d(x^*_{\mathrm{b, n}})$  (left), $r(x^*_{\mathrm{b, 
	n}})$ (middle) and cumulative time (right).}
	\label{fig:push4}
\end{figure}

For ``push4'', the location of the robust minimum is the same in that the
robot hand starts in the bottom-right, pushing to the top-left. Thus $x_r^* =
(5, -5, -0.79, 25.7)$ when using $\alpha = 0.1$.  Here $g(x_r^*, 0.1) = 1.64$
compared to $g((-5, 5, -0.79, 25.7), 0.1) = 2.71$ by pushing to the squared
target location. We again compared the methods from Section \ref{sec:empcomp}
with 1,000 MC iterations, but this time with an initial LHS of size $n_0 = 20$
and eighty acquisitions ($N=100$). Results are presented in Figure
\ref{fig:push4}. Note that increasing the dimension makes every method do
worse, but relative comparisons between methods are similar. Here {\tt
StableOPT}'s performance is better, on par with REI variations, again modulo
computing time (right panel).

\begin{figure}[ht!]
	\centering
	\includegraphics[width=5.8cm,trim=25 15 0 30]{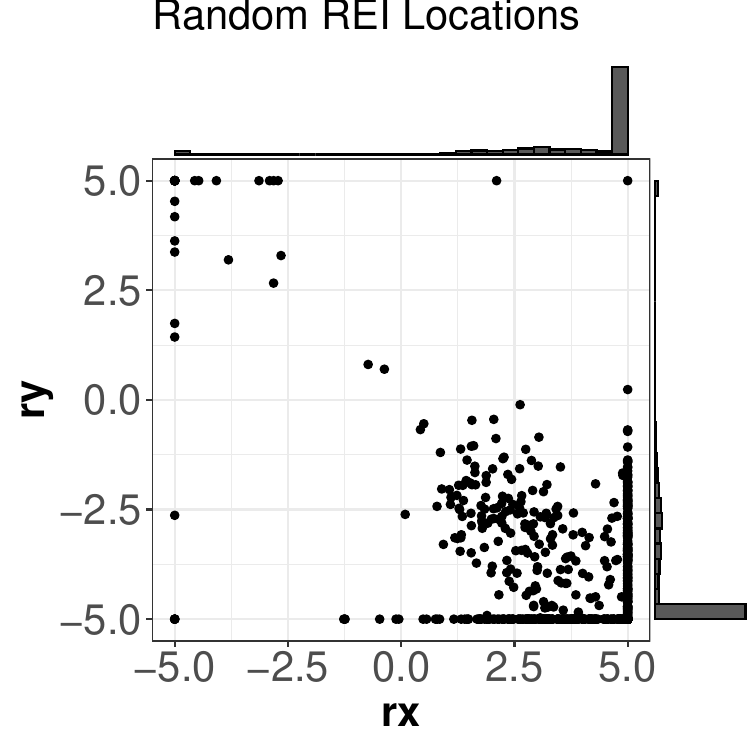}
	\includegraphics[width=5.8cm,trim=25 15 0 30]{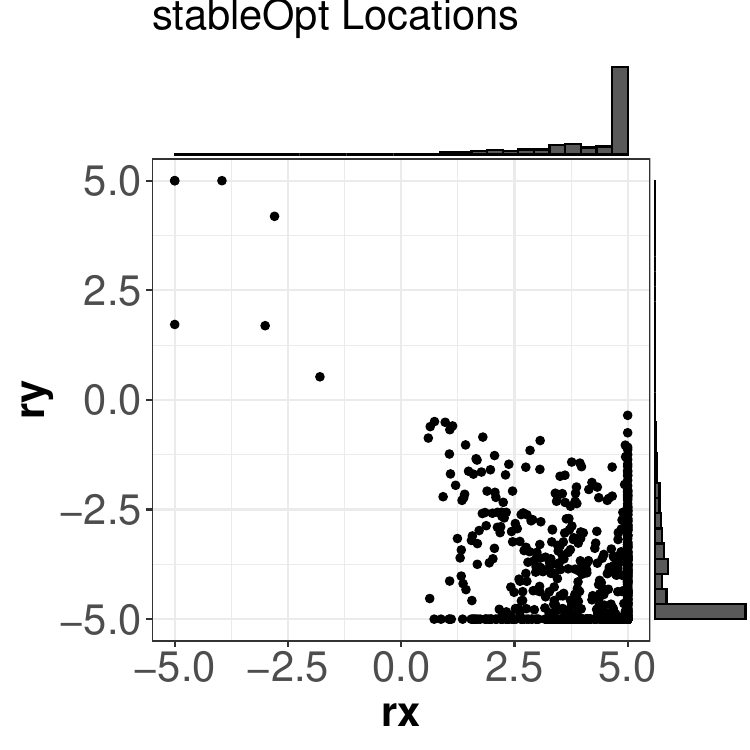}
	\includegraphics[width=5.8cm,trim=25 15 0 30]{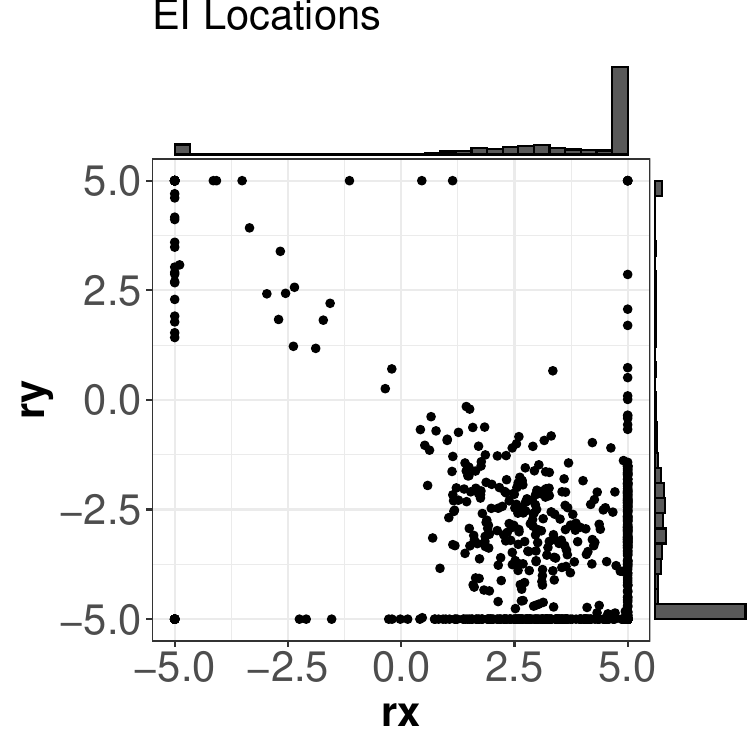}
	\caption{Distribution of $r_x$ and $r_y$ from $x^*_{\mathrm{bear}}$ for the 
	push4 problem after the final acquisition for random REI using $\alpha = 
	0.1$ (left), {\tt StableOPT} (middle), and EI (right).}
	\label{fig:push4_hists}
\end{figure}

Figure \ref{fig:push4_hists} demonstrates the added difficulty
``push4'', as indicated by additional spread in the optimal $(r_x,
r_y)$ compared to Figure \ref{fig:push3_hists}. This is because a slightly
misspecified starting location can be compensated for by adjusting the hand
angle.  Incorporating that angle introduces a slew of local minima at each
$(r_x, r_y)$. For example, if we set $r_x = 3$ rather than 5 as is optimal,
and check the regret for ``push3'' and ``push4'', we get $1.1$ and $0.7$,
respectively, by changing $r_\theta$ to $-0.69$. We may also adjust $t_r =
24.2$ in both cases to account for the \blu{hand being} closer to the box.
This phenomenon is not unique to the example we chose.  Any slight
misspecification of $r_x$ and $r_y$ \blu{can be} compensated for with a
commensurate change in $r_\theta$.

\blu{To summarize, in contrast to the synthetic examples in Section
\ref{sec:empirical}, the results presented here were obtained from a real
physics-based simulation. In the 3d variation, REI performs better than all
other comparators by finding the minimum in fewer evaluations and its solutions
settle on lower distances from the true input location of the robust minimum.
In 4d, it's a closer call: REI performs at least as well as every other
non-conventional BO method. The {\tt stapleOPT} comparator is competitive,
modulo trade-offs previously discussed.  However, we note that it was not
nearly as competitive on synthetic examples.}

\section{Discussion} 
\label{sec:conc}

We introduced a robust expected improvement (REI) criteria for Bayesian
optimization (BO) that provides good experimental designs for targeting robust
minima. \blu{REI is doubly robust, in a sense, because it is able to
accomplish this feat even in cases where one does not know, {\em a priori}, the
``proper'' level of robustness, $\alpha$, to accommodate.} In fact, we have
shown that in some cases, not fixing the $\alpha$ robustness level beforehand
leads to faster discovery of the robust minimum.  BO methods, e.g., ordinary
EI, miss the robust target entirely.  However, by blending EI with an
(surrogate modeled) adversary, we are able to get the best of both worlds.
Even when designs are not targeted to find a robust minimum, the adversarial
surrogate can be applied post hoc to similar effect.

\blu{Despite this good empirical performance, the idea of robustness
undermines one of the key assumptions of GP regression: stationarity. If one
optimum is peaked and another is shallow, then that surface is technically
nonstationary. When there are multiple such regimes it can be difficult to pin
down lengthscale(s) $\theta$, globally in the input space. \blu{Furthermore,
$Y^{\alpha}(x)$ can include flat regions which are hard to model with the
standard, stationary GP.} The underlying REI acquisition scheme may perform
even better when equipped with a non-stationary surrogate model such as the
treed GP \citep{Gramacy2007} or a deep GP \citep{Damianou2013, Sauer2022}.  We
surmise that non-GP-based surrogates, e.g., neural networks
\citep{Shahriari2020}, support vector machines \citep{Shi2020} or random
forests \citep{Dasari2019} could be substituted
\blu{for a GP. Likewise, other acquisition functions can be applied},
such as upper-confidence bound \citep{srinivas2009gaussian} or Thompson
sampling \citep{thompson1933likelihood}, without a fundamental change to the
underlying \blu{adversarial BO methodology}.}

Throughout we assumed deterministic $f(x)$.  Noise can be accommodated by
estimating a nugget parameter.  The idea of robustness can be extended to
protect against output noise \citep{Beland2017}, in addition to the ``input
uncertainty'' regime we studied here. In that setting, it can be helpful to
entertain a heteroskedastic (GP) surrogate
\citep{binois2018practical,binois2018replication, R-hetGP} when the noise
level is changing in the input space.   \blu{We also worked in relatively
modest input dimension, with examples on $d \leq 6$.  The challenge in working
in even higher dimension is two-fold: one is the practical aspect of having a
test problem in higher dimension that benefits from a robust approach.  The
largest real-world example that we could find was the robot pushing problem in
4d.  For our 6d work in the supplement we extended the Rosenbrock example from
Section \ref{sec:empirical}.  The second challenge involves surrogate modeling
in higher input dimension, where GPs are known to be data hungry.
Yet the whole point of BO is to limit collecting of expensive runs.  We
are optimistic that recent ideas from the frontier of high dimensional BO
would port well to our adversarial framework \citep{eriksson2020scalable}.}

\blu{Considering that REI/REGO is a multi-step process, and one clearly more
involved than ordinary EI/EGO, one may wonder precisely where added value is.
The theory for EI/EGO
\citep[e.g.,][]{bull2011convergence,snoek2012practical,bect2016supermartingale}
says that, among other things, eventually you will explore everywhere: an
already high level of robustness. That same theory says that EI is optimal for
the next acquisition for $f_n$, but says nothing futher about the sequence of
acquisitions. One can only ``hope'' that, with a limited budget, EI puts early
runs in the right places. That this happens has been illustrated in practice,
across many examples and variations, but there is no theory for it. By
analogy, REI is optimal for the next acquisition via $\hat{f}_n^\alpha$,
eventually spreading runs everywhere, but one can only ``hope'' for desirable
behavior in the short run, for limited budgets: less seduced by sharp, local
mimima but no less exploitative than EI. We illustrated that REI performs
desirably on a variety of examples in Sections
\ref{sec:empirical}--\ref{sec:rob}.}

\blu{Like EI/EGO, a theory that guarantees good performance for REI/REGO,
under regularity conditions that don't preclude realistic application (e.g.,
stationarity, known hyperparameterization, etc.), remains illusive.  Both will
eventually explore everywhere, but that is an un-inspiring notion of
robustness.  Perhaps someone smarter than us will come along to fix that.
What's important is ``good progress'' before reaching ``eventually''.  This is
where BO really shines.  Many of the robust methods from the mathematical
programming literature -- like those cited in Sections \ref{sec:intro} and
\ref{sec:implement} -- come with attractive theoretical results, at least
superficially, like convergence guarantees.  However, their empirical
performance isn't competitive under strict evaluation budgets that are common
in BO contexts.  Perhaps one pays the price for good theory, at least in this
context, with poorer performance in practical settings.}

\subsection*{Funding}
This work was supported by the U.S. Department of Energy, Office of Science, 
Office of Advanced Scientific Computing Research and Office of High Energy 
Physics, Scientific Discovery through Advanced Computing (SciDAC) program 
under Award Number 0000231018.

%

\bibliographystyle{jasa}
\bibliography{references}

\begin{appendices}

\section{REI with estimated lengthscale}
\label{app:esttheta}

\blu{Our experiments in Sections \ref{sec:empcomp} and in Section \ref{sec:rob} 
used fixed lengthscales $\theta$ to control MC variability.   We did that so
that the focus of those experiments to be on the acquisition criteria behind
the BO algorithms, not on the surrogate modeling details -- which are shared
identically among all the methods.  However, it is worth wondering how those
results might change when lengthscales are re-estimated after each
acquisition.}
\begin{figure}[ht!]
	\centering
	\includegraphics[width=8.75cm,trim=0 20 0 50]{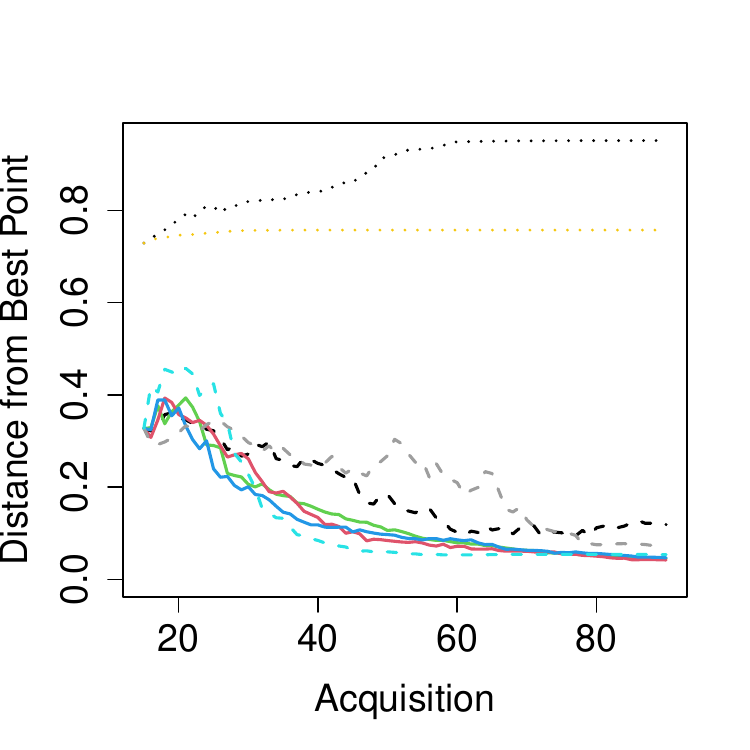}
	\includegraphics[width=8.75cm,trim=0 20 0 50]{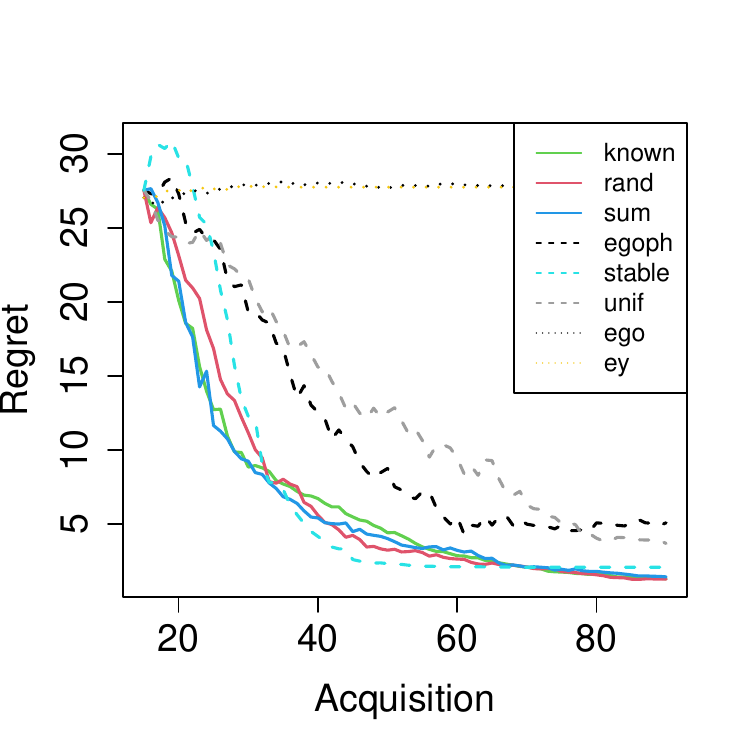}
	\caption{\blu{Results for 2d Bertsimas function with $\hat{\theta}$ 
	estimated via MLE: $d(x^*_{\mathrm{b, n}})$ (left) 
	and $r(x^*_{\mathrm{b, n}})$ (right).}}
	\label{fig:bertsima_mle}
\end{figure}
\blu{The results in Figure \ref{fig:bertsima_mle} show the Bertsimas function (from Section \ref{sec:empcomp})
with $\alpha=0.15$ in each dimension, akin to Figure \ref{fig:bertsima2d}.
Each $\hat{\theta}$ is calculated by numerically maximizing the likelihood
(i.e., MLE) of the multivariate normal log likelihood for the surrogate, and the
adversarial surrogate as necessary.   The results are very similar to the
fixed-$\theta$ case, but they are noisier due to the estimation risk in
inferring these additional hyperparamers. In particular, although REI methods
perform relatively worse compared to fixed $\theta$ case in Figure
\ref{fig:bertsima2d}, they are still winning in comparison to the other
methods.}

\section{REI in higher input dimension}
\label{app:highdim}

\blu{Figure \ref{fig:rosenbrock_6d} shows the results for the 6d Rosenbrock 
function, continuing from Figure \ref{fig:rosenbrock2d} in Section
\ref{sec:empirical}. We use the same setup as the 4d problem, described
therein, in particular with $\alpha=0.1$ for every input coordinate. For the
initial setup of the 6d problem, we included a point within 0.02 of the
global, peaked minimum in each dimension. This is not necessary in general,
however in this high dimensional space it is very difficult to locate the
global, spiky optimum.  By nudging all solvers toward discovering this area,
we are better able showcase the merits of our robust solution, which are
designed {\em not} to be fooled by the spiky solutions.  
\begin{figure}[ht!]
	\centering
	\includegraphics[width=8.75cm,trim=0 20 0 40]{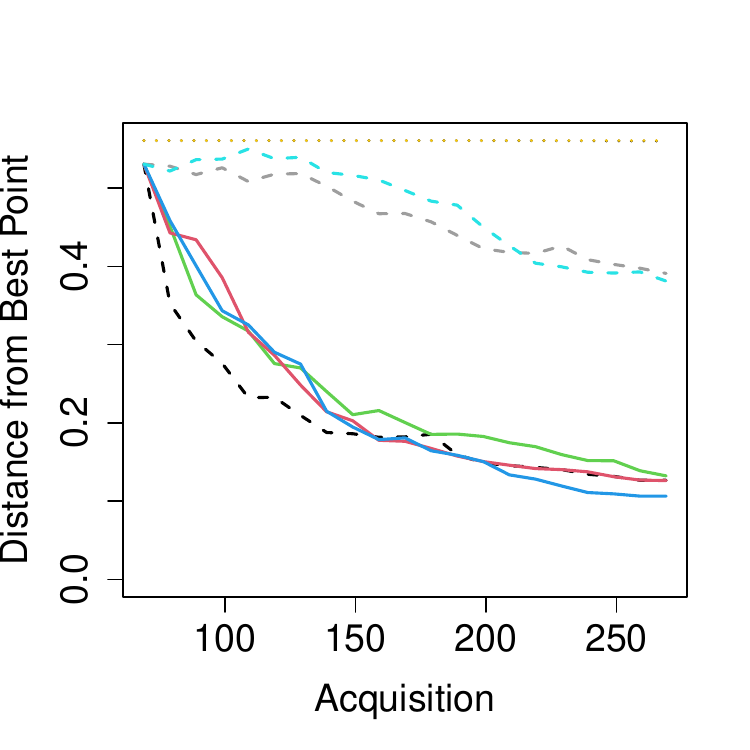}
	\includegraphics[width=8.75cm,trim=0 20 0 40]{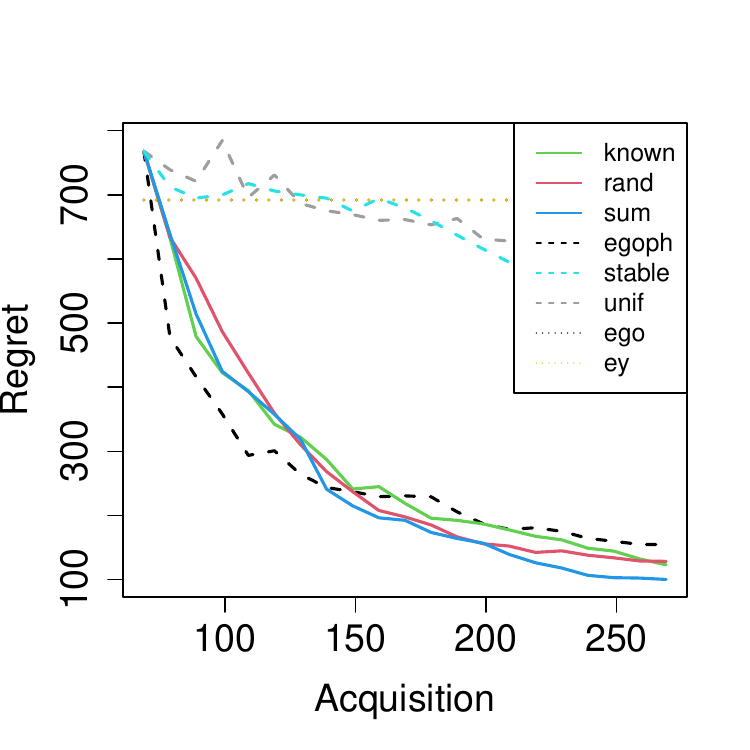}
	\caption{\blu{The results for the 6d Rosenbrock function: 
	$d(x^*_{\mathrm{b, n}})$ (left) and $r(x^*_{\mathrm{b, n}})$ (right).}}
	\label{fig:rosenbrock_6d}
\end{figure}
Adding this additional design point represents a ``temptation'' for the
conventional, non-robust BO methods, drawing them towards the global minimum
over the robust minimum. Without it, the non-robust methods rarely find the
global minimum since the surface is so peaked and thus those methods often
report the robust minimum erroneously. We see this as short-circuiting the
outcome of a much more exhaustive search.  I.e., illustrating how each of the
methods would perform in the long run, wherein eventually the non-robust
methods would be enticed by the global minimum and that would lead to worse
acquisitions. Under this setup, and pretty much identically to the 4d example
in Section \ref{sec:empirical}/Figure \ref{fig:rosenbrock2d}, the REI
methods are performing the best by both metrics: they are finding the
robust minimum faster than the other methods.}

\end{appendices}

\end{document}